\documentclass[sigconf]{acmart}

\AtBeginDocument{%
  \providecommand\BibTeX{{%
    \normalfont B\kern-0.5em{\scshape i\kern-0.25em b}\kern-0.8em\TeX}}}

\copyrightyear{2023}
\acmYear{2023}
\setcopyright{rightsretained}
\acmConference[KDD '23]{Proceedings of the 29th ACM SIGKDD Conference on Knowledge Discovery and Data Mining}{August 6--10, 2023}{Long Beach, CA, USA}
\acmBooktitle{Proceedings of the 29th ACM SIGKDD Conference on Knowledge Discovery and Data Mining (KDD '23), August 6--10, 2023, Long Beach, CA, USA}
\acmDOI{10.1145/3580305.3599933}
\acmISBN{979-8-4007-0103-0/23/08}

\usepackage{algorithmic}
\usepackage[ruled]{algorithm2e}

\usepackage{graphicx}
\usepackage{spverbatim}
\usepackage{pythonhighlight}
\usepackage{rotating}
\usepackage{colortbl}

\usepackage{xspace}

\newcommand\newcontent[1]{#1}

\begin{document}

\title{Yggdrasil Decision Forests:\\A Fast and Extensible Decision Forests Library}

\author{Mathieu Guillame-Bert}
\email{gbm@google.com}
\affiliation{%
  \institution{Google}
  \country{Switzerland}
}

\author{Sebastian Bruch}
\email{sbruch@acm.org}
\affiliation{%
  \institution{Pinecone}
  \country{United States}
}

\author{Richard Stotz}
\email{richardstotz@google.com}
\affiliation{%
  \institution{Google}
  \country{Switzerland}
}

\author{Jan Pfeifer}
\email{janpf@google.com}
\affiliation{%
  \institution{Google}
  \country{Switzerland}
}

\renewcommand{\shortauthors}{Guillame-Bert et al.}

\begin{abstract}
Yggdrasil Decision Forests is a library for the training, serving and interpretation of decision forest models, targeted both at research and production work, implemented in C++, and available in C++, command line interface, Python (under the name TensorFlow Decision Forests), JavaScript, Go, and Google Sheets (under the name Simple ML for Sheets). The library has been developed organically since 2018 following a set of four design principles applicable to machine learning libraries and frameworks: simplicity of use, safety of use, modularity and high-level abstraction, and integration with other machine learning libraries. In this paper, we describe those principles in detail and present how they have been used to guide the design of the library. We then showcase the use of our library on a set of classical machine learning problems. Finally, we report a benchmark comparing our library to related solutions.
\end{abstract}




\maketitle

\section{Introduction}

\begin{figure}[h!]
\centering
\includegraphics[scale=0.1]{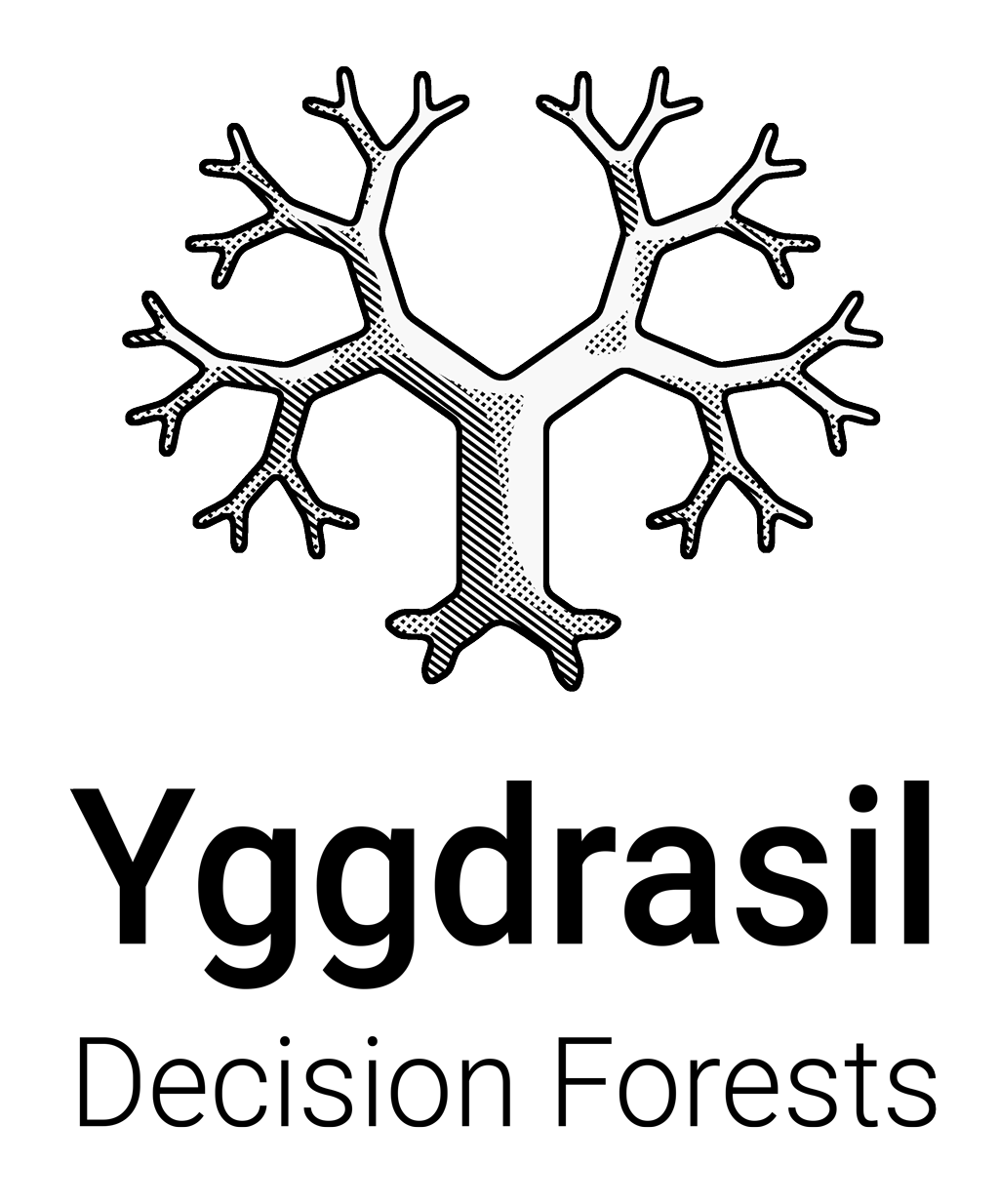}
\caption{Logo of the YDF library. Yggdrasil is a mythological holy tree in the Nordic culture. While not as ambious as the Yggdrasil tree, our library aims to reach and support all relevant domains of machine learning.}\label{fig:logo}
\end{figure}

This work introduces Yggdrasil Decision Forests (YDF), a library for the training, serving, and interpretation of decision forests\footnote[2]{Yggdrasil Decision Forests, whose logo is illustrated in Figure~\ref{fig:logo}, is available at \href{https://github.com/google/yggdrasil-decision-forests}{https://github.com/google/yggdrasil-decision-forests}.}. Decision forests are a rich class of Machine Learning (ML) models that utilize decision trees as weak learners to learn powerful functions. Decision forests stand out for their ease of use due to the relatively small number of hyper-parameters that enjoy intuitive default values and stability; their competitive and often superior performance on tabular data; their native support for numerical and categorical features without the need for preprocessing; their robustness to noisy data; their sample efficiency; and, their lightweight training and fast inference. Popular decision forests learning algorithms include Random Forests~\cite{breiman_2001}, Gradient Boosted Decision Trees~\cite{friedman_2001}, AdaBoost~\cite{schapire_2013}, and popular decision tree learning algorithms include C4.5~\cite{quinlan_1994}, and CART~\cite{breiman_1984}.

In spite of the advances in other areas of ML and the growth of neural networks in particular over the past decade, decision forests remain competitive solutions to many real-world problems and, as a technology, are integral to many production systems to date, as evidenced by the many open-source libraries~\cite{ke_2017_lightgbm,chen_2016}. As such, software libraries that facilitate the use of decision forests are central to software engineering in industry as well as academic research, with the design choices behind them having significant ramifications for research and development. While we acknowledge that this is a challenge faced by software libraries in general, it is particularly more acute in ML due to its many moving parts and the sheer volume of novel methods and algorithms that make up a training or inference algorithm. If the library proves rigid and hard to extend in the face of rapid and often unforeseen developments in the literature, by continually building atop that stack, developers and researchers alike run the risk of creating overly complex and hard-to-maintain software which may ultimately constrain product quality and the scope of research.



With that in mind, we developed YDF based on four design principles: simplicity of use; safety of use; modularity and high-level abstraction; and compatibility and integration with other ML libraries. These guidelines have proved consequential for the development of YDF, which is what motivated us to share them with the community for the benefit of software engineers and researchers. We discuss the four pillars of YDF in Section~\ref{sec:principles}. To explain how these guidelines determined the design choices in the development of YDF, we discuss its architecture in detail in Section~\ref{sec:learner_and_models}. In Section~\ref{sec:usage_examples}, we present an application of YDF to classical ML problems, followed by a complete benchmark to compare YDF with existing solutions in Section~\ref{sec:benchmarks}. Section~\ref{sec:conclusion} concludes this work.


YDF is used in several internal products with an estimated number of predictions in the tens of millions per second. YDF is integrated with TensorFlow~\cite{tensorflow_2015}, making it easy for users to integrate it and compare it to other machine learning algorithms, such as neural networks. This is particularly helpful as YDF is integrated in our internal AutoML systems, and has proven to be successful compared to deep models on certain tasks: YDF is the preferred solution in approximately a quarter of real-world use cases, with tens of thousands of models trained every day.

\section{Four design principles}
\label{sec:principles}

\subsection{Simplicity of use}

\emph{Simplicity of use} refers to the ease and efficiency with which a typical user of the library can deploy and maintain a solution. This is an increasingly difficult objective because, with the democratization of ML, the users of ML libraries now include not just researchers and experienced software engineers, but also a diverse population of non-ML researchers, students, and hobbyists among others. In addition to standard software library development best practices, therefore, simplicity in the context of an ML library entails the following properties:

\begin{description}


\item[High-level interactions and messages:] The API, documentation, error messages, and results (such as model evaluation reports), should communicate with enough abstraction that is easily digested by the user but that nonetheless provides enough context for troubleshooting. This often means communicating ideas at a high and intuitive level with as much detail as necessary to allow the user to build a mental model of the library and its workflows. In addition, error messages must provide directions and guidelines to resolve the underlying issues. To illustrate this point, we write in Table~\ref{tab:error_message} two messages that describe the same underlying error from a supervised binary classification pipeline.

\item[Sensible default values and behavior:]

The library should rely on meaningful and documented default parameters and behaviors. Those default values and behaviors should be explicit to and adjustable by the user so that they can be tailored to their specific use case. For instance, while the quality of a model can often be improved by optimizing the hyper-parameters of the learning algorithm, some hyper-parameter values give sufficiently satisfactory results in most cases (e.g., a shrinkage rate of 0.1 is reasonable for most cases in gradient boosted trees learning). As another example, YDF can automatically determine the characteristics and semantic (e.g., numerical vs. categorical) of an input feature---as detailed in Section~\ref{sec:automated_feature_injestion}. The output of this automated system is then presented to the user, which gives the user an opportunity to rectify an incorrectly determined type or to modify it arbitrarily.


In practice, YDF takes this rule one step further by adopting the following philosophy: Any operation that can be automated should be automated, the user should be made aware of the automation, and should be given control over it. This is summarized in YDF's motto: \emph{``With only five lines of configuration, you can produce a functional, competitive, trained and tuned, fully evaluated and analysed machine learning model. With four more lines of configuration, you can compare this model to any other machine learning model. With three more lines, you can deploy your model in production.''}.

\item[Clarity and transparency:] The user should have an accurate mental model of what the library does. To that end, the library must concretely define its concepts and terminology, and must accurately document metrics and algorithms with citations to the literature where appropriate. The library should explicitly note any heuristic or approximation that it uses for efficiency reasons on large datasets (e.g., evaluation on a random subset of the dataset). Reports should be self-contained, readable, and exhaustive.
\end{description}

\begin{table*}
\caption{Example of (a) poor and (b) well-written messages for an error.}
\label{tab:error_message}
\noindent\fbox{
\parbox{\textwidth}{
(a) \texttt{\small
Invalid tensor shape name="internal\_tensor\_1" shape={[}None, 4{]},
dtype=int64, {[}large stack...{]}}
}
}
\noindent\fbox{
\parbox{\textwidth}{
(b) \texttt{\small
Binary classification training (task=BINARY\_CLASSIFICATION) requires 
a training dataset with a label having 2 classes, however, 4 classe(s) 
were found in the label column "color". Those 4 classe(s) are {[}blue, 
red, green, yellow{]}. Possible solutions: (1) Use a training dataset 
with two classes, or (2) use a learning algorithm that supports 
single-class or multi-class classification e.g. learner='RANDOM\_FOREST'}
}
}
\end{table*}

\subsection{Safety of use}

Applied ML is rather unusual in that errors can lead to suboptimal yet entirely functional models! As an obvious example, tuning the hyper-parameters of a model on a held-out test dataset can lead to great offline but poor live model performance. This effect can be hard to distinguish from the impact of a distributional shift. Other common ML mishaps include modeling features according to an incorrect semantic (e.g., numerical features interpreted as categorical, thereby preventing the model from using order and scale information), or comparing trained models without accounting for the training and evaluation uncertainties.

The \emph{safety of use} principle aims to reduce the likelihood for both experienced and inexperienced users to introduce such errors and increase the chances of catching them during development. For an ML library this entails the following:

\begin{description}
\item[Warning and error messages]: Just as a compiler warns the user of potential mistakes, an ML library should look for possible errors and alert the user accordingly. When error seems likely or the impact of a potential error catastrophic, the error should interrupt the operation by default, with an option to ignore it explicitly. When error seems less likely, a non-interrupting warning will do instead.

Furthermore, the note on \emph{high-level interactions and messages} stated in the previous section applies to these warnings as well. For instance, the training of a multi-class classification model on a label that looks like a numerical value (i.e., with a large number of unique values), the error could state that: \texttt{\small The classification label column ``revenue'' looks like a regression column (4,123 unique values or 50,000 examples, 99\% of the values look like numbers). Solutions: (1) Configure the training as a regression with task=REGRESSION, or (2) disable the error with disable\_error.classification\_ look\_like\_regression=true}.

\item[Easily accessible, correct methods]: The library should make it easy to (automatically) execute what is considered ML best practices. Explanations of those practices should be well documented and easily available to the user. For example, model evaluation should contain confidence bounds with a sufficiently detailed description of how they are computed (e.g., bootstrapping, closed form solution, approximation) and how they may be used. Similarly, model comparison should include the results of appropriate statistical tests.
\end{description}

\subsection{Modularity and high level abstraction}

Modularity is a well-understood but informal blueprint for providing adaptability in software design~\cite{sullivan_2001}. In YDF, modularity implies that any sufficiently-complex part of the code may be understood as an independent module, with the interface between the various modules relying on clearly and concretely defined high-level concepts that do not expose their internals, and that are extensible and interchangeable. The initial version of every module can be as simple and generic as possible, even at the expense of execution efficiency, to facilitate readability.

At the start of development, modularity incurs a development cost and may therefore seem unwarranted. In fact, overly generic and slow code can be counterproductive. In YDF, however, we observed that modularity brought about many advantages as we elaborate shortly. As the library grows with newly published techniques, some modules become overly complex or inefficient. In cases like that, modularity allows for the re-writing of a specific module in isolation without having to understand the library in its entirety. It also allows recycling unit- and integration-tests of the previous version of the module. A re-write may often involve breaking up a single module into sub-modular parts.

Consider, as an example, the \emph{splitter} routines that are at the core of every decision tree learning algorithm and are responsible for selecting split conditions for an intermediate node in the tree. In YDF, the initial splitter implementation was a single module handling only numerical features on classification labels. The splitter was implemented using an ``in-sorting'' approach (i.e., sorting feature values for each node) making it simple to implement and test, usable for both deep and shallow trees, but slower than more advanced or specialized splitters. A few other splitters supporting other common feature types were later added as separate modules.

As support for other types of features (e.g., categorical, pre-sorted numerical, categorical-set~\cite{guillame_bert_2020_catset}, with or without missing values), other types of labels (e.g., regression, ranking), parameters (e.g., divide and conquer growth, global growth~\cite{shi_2007}, oblique splits~\cite{tomita_2020}), and shape of trees (e.g., shallow, deep) became necessary, the splitters were refactored and sub-modularized. YDF splitter code is now organized into three types of modules responsible for label type, feature type, and the specific splitting logic, with the resulting organization favoring code reuse and reducing the cost of extension and maintenance. For example, the module handling binary classification (label type) is used for all feature types, and the module handling categorical features (feature type) is used for all label types. This new design incorporated the engineering experience acquired during the implementation of the first generation of splitters.

Modularity also allows for the cohabitation of both generic-and-slow and specialized-and-fast code, where the initial simple modules serve as the ground truth in unit testing of specialized or optimized modules. For example, as mentioned above, the first YDF splitters used a simple in-sorting approach. Later, more efficient but complex splitters (e.g., pre-sorting approach, distributed approach) used the in-sorting approach as unit tests. In the case of deep trees (e.g., trees trained by the Random Forest learning algorithm), in-sorting is sometimes more efficient than pre-sorting. The modularity thus allows YDF to dynamically choose the most efficient splitter implementation for each node.


\subsection{Integration with other ML libraries}

ML libraries and frameworks should easily interact with each other and allow compositions. The possibility of \emph{interaction} between libraries reduces the risk of the ``framework trapping effect,'' in which the user is limited to the methods available in the library they are most familiar with, or on which the project relies, possibly missing some more suitable methods available in other libraries, resulting in sub-optimal solutions. For example, while R~\cite{R_2022} contains a rich variety of data mining and decision forest libraries, it is not trivial for TensorFlow~\cite{tensorflow_2015} users to use them.

\emph{Composability} is important for both research and advanced production work. For example, the composition of decision forests and neural networks can lead to improvements in model quality~\cite{bruch_2020_differentiable,li_2019_combine,guillame_bert_2020_catset,guolin_2019_deep_gbm}. But neural network libraries often show poor efficiency when executing the branching algorithms used to train decision forests, making the composability of neural networks and decision forests libraries a prerequisite for such hybrid research.

\section{Structure of YDF}

\begin{figure}[t]
\centering
\includegraphics{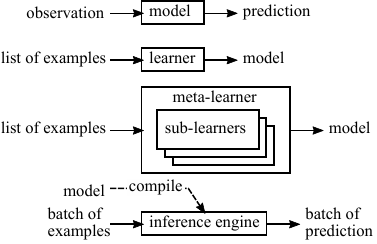}
\caption{High level modules in YDF.}\label{fig:structure}
\end{figure}

In this section, we present the design decisions behind YDF and show the role the principles of Section~\ref{sec:principles} played in the formation of these decisions. Figure~\ref{fig:structure} depicts the high-level structure of YDF. We explain each of the components in this figure in the remainder of this section.

\subsection{\textsc{Learner}-\textsc{Model} abstraction}
\label{sec:learner_and_models}

YDF's inter-module communication and user API relies on two model-agnostic ML concepts: \textsc{Learner}s and \textsc{Model}s. A \textsc{Model} is a function, in the mathematical sense, that takes an observation as input, and returns a prediction.
A \textsc{Learner} is a function that takes a list of examples as input and returns a \textsc{Model}.
An \textsc{example} is a couple \{\textsc{observation}, \textsc{label}\}.


The \textsc{Learner}-\textsc{Model} abstraction is simple but generic enough for the integration of any new learning algorithm because it makes few assumptions about how the learning algorithm or model work internally. \textsc{Learner}s, for example, are not required to rely on stochastic gradient descent for optimization.

The \textsc{Learner}-\textsc{Model} abstraction is commonly used in the non-parametric learning R~\cite{R_2022} packages~\cite{wiener_2002,therneau_2002,wright_2017}. In contrast, in Python ML libraries, we more routinely encounter the \textsc{Estimator-Predictor} paradigm. For example, in scikit-learn~\cite{sklearn_api} both the training and inference logic are encoded into the same object using the \emph{model.fit} and \emph{model.predict} functions. A similar design choice exists in XGBoost~\cite{chen_2016} and TensorFlow~\cite{tensorflow_2015}. Note that, for consistency, the port of YDF in TensorFlow, called \emph{TensorFlow Decision Forests}, uses the \textsc{Estimator-Predictor} abstraction.

We argue that the distinction between \textsc{Learner}s and \textsc{Model}s allows for the separation of training and inference logic (the inference logic is generally simpler than training) as well as code reuse (i.e., different \textsc{Learner}s can train the same type of \textsc{Model}, and a given \textsc{Learner} can produce different types of \textsc{Model}s). For example, \citet{breiman_1984} and \citet{guillame_bert_2018} are two algorithms to train Random Forest models. While the algorithms are different in the way they learn random forests, the models they produce have a similar structure, and the same post-training Random Forest-specific tools are applicable to both \textsc{Learner}'s outputs.
Finally, the separation of the learning and inference logic facilitates the development of technology-agnostic tools such as hyper-parameter tuners, cross-validation learner evaluators, model ensemblers, feature selection algorithms, and model agnostic interpreters.

To illustrate the benefit of separating the learning and inference logic for library integration and efficiency, consider the following example.
Suppose a \textsc{Learner} trains a linear and a decision tree \textsc{Model} using two separate external libraries and returns the best performing one.
The \textsc{Model} returned by the learner is either a linear model \emph{or} a decision tree, compatible with the tooling and the respective external library.
Deploying the model to a production service so as to generate predictions only requires loading the inference logic of one of the models.
By comparison, if the learning and inference logic are packed into the same object, this object is not directly compatible with the external libraries. Finally, loading the model in a production setting to generate predictions requires loading (or at least making accessible) the inference and training logic of both models and the model selection logic.

In YDF, \textsc{Models} and \textsc{Learners} are implemented by deriving abstract model and learner C++ classes respectively. This abstraction is independent of the task at hand: YDF includes \textsc{Learners} with support for classification, regression, ranking, and uplifting tasks. The abstract classes expose various additional functionality common to many learners and models, such as (de-)serialization, determining variable importance, and human-readable summary.

A new \textsc{Learner} can be integrated into YDF using a C++ registration mechanism (e.g., \texttt{REGISTER\_AbstractLearner(MyLearner)})---see Section~\ref{sec:modularity} for details on the YDF registration mechanism. YDF comes with a few built-in \textsc{Learner}s such as CART~\cite{breiman_1984_cart}, Random Forest~\cite{breiman_2001} and Gradient Boosted Trees~\cite{friedman_2001}, as well as \emph{meta}-learners that we will describe in Section~\ref{sec:meta_learners}. Additionally, YDF offers learners that are effectively wrappers to other ML libraries.


\subsection{Meta-learners}
\label{sec:meta_learners}

One of the interesting properties of the \textsc{Learner}-\textsc{Model} abstraction is that it allows for the composition of algorithms. To illustrate this point, consider a hyper-parameter tuner which is code that is responsible for finding the optimal hyper-parameters of a learner. It turns out that a hyper-parameter tuner can itself also be thought of as a \textsc{Learner}: It returns a model trained with a base \textsc{Learner} but using the optimal hyper-parameter values. To make matters more interesting, the method used by the hyper-parameter tuner to assess the optimality of candidate hyper-parameters (e.g., cross-validation, train-validation) is itself a hyper-parameter of the hyper-parameter tuner! We call all such \textsc{Learner}s that use another or multiple other \textsc{Learner}s, \textsc{Meta-Learner}s. 

Other \textsc{Meta-Learner}s include: the ``calibrator'' which calibrates the predictions of a \textsc{Model}; the ``ensembler'' which ensembles a set of \textsc{Model}s; and the ``feature selector'' which determines the optimal subset of input features for a \textsc{Learner} on a given dataset. 

\begin{figure}[t]
\centering
\includegraphics{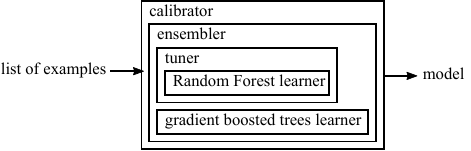}
\caption{Representation of the three imbricated \textsc{Meta-Learners}. }\label{fig:example_meta_learner}
\end{figure}

\textsc{Meta-Learners} too can be composed together. Figure~\ref{fig:example_meta_learner} shows an example of a calibrator \textsc{Meta-Learner} containing an ensembler, which itself contains both a hyper-parameter tuner optimising a Random Forest \textsc{Learner}, and a vanilla (i.e., without hyper-parameter tuning) Gradient Boosted Tree \textsc{Learner}.

\subsection{Model validation}

By default, YDF \textsc{Learners} do not take validation datasets as input. Instead, if the learning algorithm requires a validation dataset (e.g., for early-stopping), it is extracted once (for train-validation) or multiple times (for cross-validation) from the training dataset by the \textsc{Learner} implementation itself. The amount of examples to extract is a hyper-parameter of the \textsc{Learner}. For applications where distribution shift is a potential issue, YDF \textsc{Learners} support an optional validation dataset as input. Each \textsc{Learner} can use this validation dataset as desired.

\subsection{Automated feature ingestion}
\label{sec:automated_feature_injestion}

One of the more important facets of an input feature is its semantic, which directly determines its mathematical properties. Using the appropriate feature semantics is critical to train good models. YDF defines several model agnostic feature semantics: \emph{Numerical features} (i.e., features with numerical semantics) are values in a continuous or discrete space with total ordering and scale significance. These generally represent quantities or counts, such as the age of a person, or the number of items in a bag. \emph{Categorical features} are values in a discreet space without order. They are generally used to represent types such as the color $red$ in the set $\{red, blue, green\}$. Other semantics include structural information like \emph{categorical sets} (i.e., where a value is a set of categories), \emph{categorical lists}, \emph{numerical lists}, \emph{numerical sets}, \emph{text}, \emph{boolean}, and \emph{hashes}.

It is worth noting that the value of a feature may be \emph{missing} or \emph{unavailable} and that, for multi-valued features such as categorical sets in particular, a missing value is semantically different from the empty set. Algorithms take great care in handling missing values and different algorithms can handle missing values differently, which is referred to as imputation strategy. The decision tree learning algorithm used in decision forest \textsc{Learner}s, for example, supports \emph{global} and \emph{local} imputation where missing values are replaced by the mean (numerical features) or most frequent (categorical feature) value estimated on the whole dataset (global) or the examples in the same tree node.

Generally speaking, the semantics of an input feature cannot be determined reliably from values or its representation. For example, the string ``2'' in a CSV file could be a numerical value, a categorical value, free text, or a numerical list with only one element. YDF uses a number of heuristics to assist the user in automatically determining feature semantics and build auxiliary data structures and metadata as required for the given feature type (such as dictionaries for categorical features).
As we noted previously, while basic heuristics often yield reasonable results, YDF insists that the user validate and optionally modify the automatically determined feature semantics.

\subsection{Modularity}
\label{sec:modularity}

YDF is organized into interchangeable modules.
Most modules are implemented with object-oriented programming inheritance, that is, by deriving and registering an abstract class.
The \textsc{Learners} and \textsc{Models}, ``inference engines'', and ``distributed computation backend'' modules are presented respectively in Sections~\ref{sec:learner_and_models}, \ref{sec:infernce_engine}, and \ref{sec:distributed_training}. Here, we describe other noteworthy modules.

\begin{description}
\item[\textsc{Reader}s] read a stream of examples. Different dataset readers support different file formats.
\item[\textsc{Writer}s] write a stream of examples as a dataset with support for many file formats.
\item[Decision tree IO] imports and exports a decision tree from and to a file stored on disk. This module is used by all models made of trees.
\item[\textsc{Splitter}s] are algorithms that find the splitting conditions in a decision tree.
\end{description}

Official modules are directly part of the YDF code base. But custom modules can be hosted outside of the YDF code base using a Bazel~\cite{bazel_2015} third-party dependency.

\subsection{Model self evaluation}
\label{sec:model_self_evaluation}
It is often essential to validate the quality of a model (i.e., to determine if quality is satisfactory) and use this information to direct model development (i.e., select the more performant model). A typical method to estimate model quality is by evaluating metrics of interest on held-out examples that are not seen by the learning algorithm. While simple, this method can be problematic and unstable in applications with a small amount of labeled data.

Other approaches to obtaining a fair estimate of model quality includes out-of-bag evaluation and cross-validation methods. In YDF, we abstract out all such model validation methods as a model \textsc{Self-Evaluation} module, leading to a powerful model-agnostic abstraction that can be utilized by \textsc{Learner}s and \textsc{Meta-Learner}s alike. For example, the feature-selector \textsc{Meta-Learner} can choose the optimal input features for a Random Forest \textsc{Model} using Out-of-bag \textsc{Self-Evaluation}.

\subsection{Inference engine}
\label{sec:infernce_engine}

The most na\"ive algorithm to compute the prediction of a decision tree is made up of a single \emph{while} loop that iterates from the root node of the tree, taking the left or right branch according to the node condition, and terminating at one of the leaves. This is shown in Appendix~\ref{alg:default_serving}.

This simple algorithm is, however, inefficient on modern CPUs due to its slow and unpredictable random memory access pattern and branching mispredictions~\cite{nima_2014}. This observation inspired a line of research to optimize tree inference by using more complex but more efficient tree traversal logic. A prominent example of tree inference algorithms is QuickScorer~\cite{lucchese_2015}. It can efficiently infer decision trees with up to $64$ nodes on a $64$-bit CPU, with the obvious caveat that it does not extend to larger trees such as those generated by the Random Forest algorithm~\cite{breiman_2001}.

In addition to the tree traversal algorithm, another factor that contributes to the latency of tree inference is the types of conditions in decision nodes. An inference algorithm that only supports one type of condition will inevitably be faster than an algorithm that supports many types. Additionally, instruction-level parallelism when available often has an outsize impact on latency. Finally, hardware accelerators (e.g. GPU, TPU, FPGA) allow for significant optimization as well.

To handle this diversity of solutions and to maximize model inference speed, but to shield the user from this complexity, YDF introduces the concept of \emph{inference engine} or \emph{engine} for short. An engine is the result of a possibly \emph{lossy} compilation of a \textsc{Model} for a specific inference algorithm. In other words, we compile a \textsc{Model} into an engine, which is chosen based on the model structure and available hardware. In this way, space, complexity, and latency can be traded off depending on which factors are important to a particular production environment. For example, when the program size is important, such as on embedded devices like Internet-of-things, and when the model is known in advance, YDF can be compiled with only the required engine.

\subsection{Splitters}
\label{sec:splitters}
Splitters are modules that find the optimal decision for a given node according to a splitting criteria. Their complexity, therefore, is tied to the number and type of features as well as the cardinality of their space of values. By default, YDF's splitters are \emph{exact} for numerical features, in the sense that numberical values are taken on face value and are in no way transformed (e.g., by discretization). This naturally leads to more candidate splits to be considered, an approach that could prove slow when the value of a feature cover a large range. This exact approach is similar to XGBoost~\cite{chen_2016} but different from LightGBM~\cite{ke_2017_lightgbm}. Like those libraries, YDF also supports \emph{approximate} splitting but discretization, leading to a significant speed-up at the cost of a potential degredation to model quality.

In addition to numerical features, YDF natively supports categorical features with exact splitting~\cite{fisher_1958_exact} (similar to LighGBM), random categorical projection~\cite{breiman_2001}, and one-hot encoding (similar to XGBoost). Finally, YDF has special support for oblique numerical splits~\cite{tomita_2020} and categorical-set splits~\cite{guillame_bert_2020_catset}.

\subsection{Distributed training}
\label{sec:distributed_training}

Distributed training is essential for training on large datasets and computationally-intensive learning algorithms such as hyper-parameter tuners. To facilitate the development of distributed algorithms, YDF defines an API with the primitives necessary for decision forest distributed training, along with a distributed training framework for common \textsc{Meta-Learner}s, all with built-in fault-tolerance.

The implementation of this API are modulable, with two particular implementations available based on gRPC and TensorFlow Parameter Server distribution strategies. But because the development and testing of distributed algorithms can be cumbersome, YDF also contains a third implementation specialized for development, debugging, and unit-testing. This implementation simulates multi-worker computation in a single process, making it easy to use breakpoints or execute the distributed algorithm step by step. How the user selects which distributed implementation to use is a single piece of configuration.

The YDF implementation of decision forests distributed training relies on both ``feature parallel'' and `example parallel'' distribution based on the work of~\citet{guillame_bert_2018}. Each \emph{training worker} is responsible for a subset of input features. Communication between workers is optimized with a delta-bit encoding and multi-round hierarchical synchronization so as to minimize the maximum network IO among workers. The type and number of features allocated to each worker is dynamically adjusted to handle fluctuation in worker availability due to concurrent execution. Workers evaluate the quality of the model on a validation dataset, and possibly trigger early stopping of the training.


\subsection{Multi API and integration with other ML frameworks}
\label{sec:other_framework_interraction}

C and C++ code is generally well supported by other languages. YDF is available in C++, on the web platform using JavaScript and WebAssembly, in Go, and Command line interface (CLI). YDF is also available in Python through TensorFlow~\cite{tensorflow_2015} under the name TensorFlow Decision Forests, making it compatible with the TensorFlow ecosystem. YDF supports NumPy~\cite{harris_2020} and Pandas~\cite{mckinney_2010} making it easy to use on small datasets in Python. YDF can read scikit-learn~\cite{sklearn_api} decision forest models. \newcontent{YDF is also available as a Google Sheets add-on under the name Simple ML for Sheets~\cite{sml_for_sheets}. This add-on makes standard modeling actions such as model training, evaluation and interpretation available without coding. It also exposes ML concepts to non-ML savvy users. For example, the task of ``predicting missing values'' is implemented using model training+inference, while the task of ``detecting abnormal values'' is implemented using cross-validation.}

It is worth noting that models and training configurations are cross-API compatible. For example, a model trained with the Python API can be run with the JavaScript API.

\subsection{Backwards compatibility and default values}
\label{sec:backward_compatibility}

YDF models are fully backwards compatible---as an anecdote, models trained in 2018 are still usable today. Additionally, the YDF training logic is deterministic: The same \textsc{Learner} on the same dataset always returns the same \textsc{Model}. This last rule may only be violated by changes in the underlying pseudo-random number generator implementation.

An important property of YDF, and a constraint we impose on the development of the library, is that hyper-parameters are backwards compatible: Running a \textsc{Learner} configured with a given set of hyper-parameters always returns the same \textsc{Model}---modulo changes to the pseudo-random number generators. This implies that default hyper-parameters cannot change and that all newer methods of learning are disabled by default. 
By construction, the default values of all hyper-parameters are set to the values recommended in the paper that introduces the algorithm or in the authors' implementation of it. For example, by default, classification Random Forest uses an attribute sampling ratio of the square root of the total number of features as recommended by Breiman~\cite{breiman_2001}.

To simplify the use of the library, particularly for users who would like to use the latest algorithms in YDF but who are not well-versed in the literature and may not understand fully the hyper-parameters involved, YDF offers a hyper-parameter template system. For example, a learner configured with the \texttt{benchmark\_rank1} parameter template will be trained with the best hyper-parameters according to our benchmark on a large number of real-world datasets. The \texttt{benchmark\_rank1@v1} template for the gradient boosted trees learner~\cite{friedman_2001} uses global tree growth~\cite{shi_2007} (i.e., best first or leaf-wise growth), sparse oblique splits~\cite{tomita_2020}, and random categorical splits~\cite{breiman_2001}. As new versions of YDF are released, those hyper-parameters can change but YDF retains version information. For example, a learner configured with the \texttt{benchmark\_rank1@v1} will be trained on the best hyper-parameters in version $1$ of this template.

\section{Usage examples}
\label{sec:usage_examples}

This section demonstrates a use case where we apply YDF for binary classification on the Adult dataset (also known as the Census Income dataset). This dataset is stored in two CSV files containing the training and test examples. Input features are either numerical or categorical, with some feature values missing.


We first highlight how one may use YDF with the CLI API to train, evaluate and analyse a classical gradient boosted trees model. This API is similar (but less verbose) than the C++ interface. We then show how to use the TensorFlow Decision Forests API \newcontent{and the Simple ML for Sheets tool} to do the same task. In all three cases, the hyper-parameters of the learner are left to their default values. In addition, the input features are not explicitly fed to YDF; instead, YDF will use all available features (excluding labels) with automated semantic detection.

\subsection{The CLI API}
\label{sec:cli_api}

The following is the CLI usage example. The resulting artefacts---dataspec, model information, model evaluation report, and model inference benchmark report---are included in Appendix~\ref{sec:artefacts_cli} for completeness.

\begin{small}
\begin{verbatim}
# Detect feature semantics, producing dataspec
infer_dataspec --dataset=csv:train.csv 
  --output=dataspec.pbtxt
  
# Print details of the inferred semantics
show_dataspec --dataspec=dataspec.pbtxt

# Configure the learner
cat <<EOF > learner.pbtxt
  task: CLASSIFICATION
  label: "income"
  learner: "GRADIENT_BOOSTED_TREES"
EOF

# Train the model
train --dataset=csv:train.csv \
  --dataspec=dataspec.pbtxt \
  --config=learner.pbtxt \
  --output=model_path

# Display information about the model
show_model --model=model_path

# Evaluate the model
evaluate --dataset=csv:test.csv \
  --model=model_path

# Generate model predictions
predict --dataset=csv:test.csv \
  --model=model_path \
  --output=csv:predictions.csv

# Benchmark the model inference speed
benchmark_inference --dataset=csv:test.csv \
  --model=model_path
\end{verbatim}
\end{small}

\subsection{The Python and Tensorflow API}

In this section, we showcase the TensorFlow Decision Forests API.

The port of YDF in TensorFlow does not use the YDF model evaluation logic demonstrated in Section~\ref{sec:cli_api}. Instead, TensorFlow native metric implementations are used for evaluation. 

\begin{small}
\begin{python}
import tensorflow_decision_forests as tfdf
import pandas as pd

# Load datasets as Pandas dataframes
train_df = pd.read_csv("train.csv")
test_df = pd.read_csv("test.csv")

# Convert the datasets to TensorFlow format
train_ds = tfdf.keras
    .pd_dataframe_to_tf_dataset(train_df,
    label="income")
test_ds = tfdf.keras
    .pd_dataframe_to_tf_dataset(test_df,
    label="income")

# Train a model
model = tfdf.keras
    .GradientBoostedTreesModel()
model.fit(train_ds)

# Summary of the model structure.
model.summary()

# Evaluate the model.
model.compile(metrics=["accuracy"])
print(model.evaluate(test_ds,
    return_dict=True))
\end{python}
\end{small}

\subsection{Simple ML for Sheets}

\begin{figure}[t]
\centering
\includegraphics[scale=0.45]{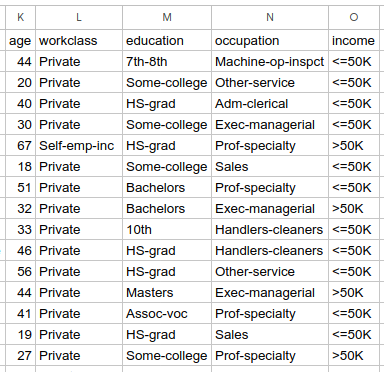}
\caption{Example of training data presented in a spreadsheet in Simple ML for Sheets.}\label{fig:sml_data}
\end{figure}

\begin{figure}[t]
\centering
\includegraphics[scale=0.45]{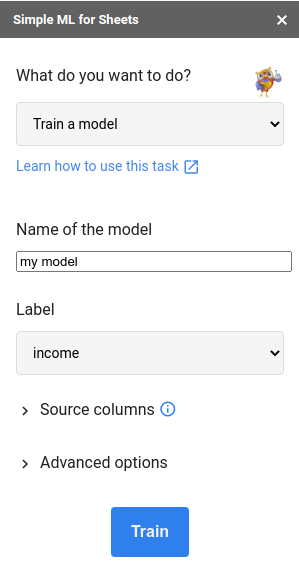}
\caption{Configuration of the ``Train a model'' task in Simple ML for Sheets.}\label{fig:sml_train}
\end{figure}

\newcontent{In this section, we showcase the Simple ML for Sheets tool. The dataset is provided as a training and testing sheets of data as shown in fig.~\ref{fig:sml_data}. The user open the training sheet, selects the ``train a model'' task and clicks on ``Train'' (fig.~\ref{fig:sml_train}). After a few seconds, a model is trained with default hyper-parameters. The user then runs the ``evaluate a model'' task and obtain an evaluation report similar to the one shown in sec.~\ref{sec:artefacts_cli_eval}. Finally, the user runs the ``Understand a model`` task and a details model report similar to the one presented in sec.~\ref{sec:artefacts_cli_dataspec}, sec.~\ref{sec:artefacts_cli_info} and fig.~\ref{fig:evaluation_report}.}

\section{Experiments}
\label{sec:benchmarks}

We compare the accuracy of YDF v0.2.5 to four popular decision forests learning libraries: XGBoost v1.5.1 (XGB)~\cite{chen_2016}, scikit-learn v1.0.2 (SKLearn)~\cite{sklearn_api}, LightGBM v3.0.0.99 (LGBM)~\cite{ke_2017_lightgbm}, and TensorFlow BoostedTrees Estimators v2.9.1 (TF BTE)~\cite{ponomareva_2017_tfb}). We also include a linear classifier (TF Linear) on 70 small tabular (binary and multi-class) classification datasets from the OpenML 
repository~\cite{vanschoren_2013_open_ml}. The list of datasets used in our evaluation appears in Appendix~\ref{sec:benchmark_dataset_stats}. The number of examples ranges from $150$ to $96,320$, with a mean of $8,647$ examples per dataset. The number of features ranges from $5$ to $1,777$, with a mean of $119$ input features per dataset.

\subsection{Learners}

For each library, we evaluate learners using both their default hyper-parameters as well as hyper-parameter values tuned using an automated tuner. The default hyper-parameters of each library might differ, except for the "number of trees" which is universally fixed to $500$. Learners that use the default hyper-parameter values are tagged with ``\emph{(default hp)}.''

As noted in Section~\ref{sec:backward_compatibility}, YDF sets the default values for all hyper-parameters to reflect the configurations in the original publication. To complement these results, we also evaluate a setting in which hyper-parameters are drawn from our benchmark configuration \texttt{benchmark\_rank1@v1}, and tag these learners with \emph{(benchmark hp')}. The definition of the \emph{default} and \texttt{benchmark\_rank1@v1} hyper-parameters are presented in appendix~\ref{sec:default_hyper_paramters}.

Tuned learners are tagged with \emph{Autotuned}. We conduct hyper-parameter tuning by aggregating results from $300$ unique random trials. Trials are scored either by log loss (noted \emph{(opt loss)}) or accuracy (noted \emph{(opt acc)}). For Random Forest models we use out-of-bag evaluation for validation, whereas for Gradient Boosted models we set aside $10$\% of the training data for validation.
The hyper-parameters tuned by each library are listed in Appendix~\ref{sec:benchmark_hyper_paramters}.

Finally, we note that Scikit-learn, XGBoost and TensorFlow BoostedTrees Estimators libraries do not offer native support for categorical features. As such, for these learners, we encode all categorical features using one-hot encoding.

\subsection{Metrics}

We evaluate all pairs of dataset and learner using a $10$-fold cross-validation protocol, where fold splits are consistent across learners to facilitate a fair comparison. Note that, the hyper-parameter tuning is applied independently on each of the $10$ folds for each library.

We measure accuracy, AUC (for binary classification), training time, and inference latency of each model. We further report the overall mean and median rank of each learner across all datasets, the number of wins or losses between pairs of learners, and the mean training and inference time of each learner.

\subsection{Computing resources}

We train each model on a single machine without using distributed training and with a $20$ threads limit. The inference of the model is evaluated with a single thread for YDF, and a number of threads selected by the library for other learners. The reported training and inference times exclude dataset reading. In aggregate, we trained a total of $1.3$ million models consisting of $840$ million trees.

\subsection{Results}

\begin{figure}[t]
\centering
\includegraphics{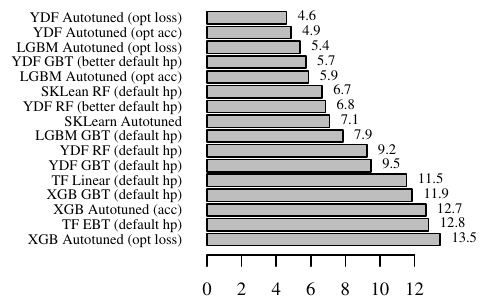} 
\caption{Mean learner ranks: The rank of the 16 learners averaged over the $70$ datasets. The smaller, the better.}\label{fig:benchmark_mean_rank}
\end{figure}

Due to space constraints, we have included the mean cross-validation accuracy of all learners on every dataset in Appendix~\ref{sec:benchmark_accuracy}, and a pairwise comparison of learners Appendix~\ref{sec:benchmark_pairwise}. Here, we present a summary of these results in Figure~\ref{fig:benchmark_mean_rank} where we render the \emph{mean rank} of each learner---equivalent to the ``mean rank'' column in the table reported in Appendix~\ref{sec:benchmark_accuracy}.
The mean rank is the average, over all datasets, of the rank of the learner (between 1 and 16) compared to other learners.
We also show in Table~\ref{tab:durations} the average training and inference duration for each learner.

\begin{table}
\caption{Training and inference duration in seconds of (un-tuned) learners, using the default hyper-parameters of the respective libraries, averaged over $10$ cross-validation folds and $70$ datasets. Learners are ordered according to their quality rank reported in Figure~\ref{fig:benchmark_mean_rank}. Individual measures for each datasets and learners are given in Appendix~\ref{sec:benchmark_training_time} and \ref{sec:benchmark_inference_time}.}
\label{tab:durations}
\center
\begin{tabular}{@{}lll@{}}
\toprule
Learner                  & \multicolumn{1}{l}{training (s)} & \multicolumn{1}{l}{inference (s)} \\ \midrule
YDF GBT (benchmark hp) & 39.99 & 0.108 \\
SKLean RF (default) & 7.01 & 0.250 \\
YDF RF (benchmark hp) & 29.86 & 0.598 \\
LGBM GBT (default) & 4.91 & 0.061 \\
YDF RF (default) & 4.41 & 0.326 \\
YDF GBT (default) & 34.79 & 0.044 \\
TF Linear (default) & 55.59 & 8.050 \\
XGB GBT (default) & 20.72 & 0.015 \\
TF EBT (default) & 212.06 & 1.949 \\
\bottomrule
\end{tabular}
\end{table}

\subsection{Observations}

On the mean accuracy computed over $70$ datasets, we make the following observations:

\begin{itemize}
    \item No one learner is always better than another learner. The largest difference is between auto-tuned YDF and XGBoost with $612$ wins and $88$ losses. The effective difference of quality between closely-ranked learners is rather small. For example, the average difference in accuracy between auto-tuned YDF and LightGBM is about $0.5$\% for a standard deviation of $1.9$\%---see Appendix~\ref{sec:benchmark_accuracy}.
    \item YDF performs better than other candidate libraries both with and without automatically tuned hyper-parameters. The YDF setting where the tuner optimizes for loss has rank $1$, and YDF with benchmark hyper-parameters has rank $4$, standing above all other non-tuned learners.
    \item Automatically tuned LightGBM comes in close second place with respect to mean rank, but ranks first in terms of the number of pairwise wins and losses.
    \item For Random Forests and Gradient Boosted Trees learners, YDF with all its features enabled---see Section~\ref{sec:backward_compatibility}---performs significantly better than YDF with the default hyper-parameters.
    \item Tuning the hyper-parameters increases the quality of YDF and LightGBM learners significantly ($+4$ and $+6$ rank change respectively).
    \item Surprisingly, scikit-learn and XGBoost without hyper-parameter tuning offer slightly better (respectively $+2$ and $+1$ rank change) results than with hyper-parameter tuning. We believe that is due to the relatively small size of the datasets and the one-hot encoding of categorical features, tuning the hyper-parameters makes the model more prone to overfitting. Note that the same hyper-parameter tuning library is used for all the learners.
    \item XGBoost and TF Boosted Trees on average yield a lower accuracy than a linear model. Gradient Boosted Trees models perform better than Random Forest models in terms of accuracy.
\end{itemize}

Moving on to the training speed of the models, we observe that LightGBM is significantly faster to train than YDF and XGBoost models for the GBT algorithm. This comparative performance can be explained by the difference in the configuration of the splitter algorithms in the three libraries. See Section~\ref{sec:splitters}. YDF is slightly faster than scikit-learn for the RF algorithm but much faster than TensorFlow based models.

On the inference speed of the models, we observe that:

\begin{itemize}
    \item Ignoring the different number of threads used, the inference speed of gradient boosted tree models trained default hyper-parameters is the fastest with XGBoost, followed by YDF, followed by LightGBM. Gradient Boosted Trees models trained with TF Estimator Boosted Trees are two orders of magnitude slower.
    \item As expected, the much larger Random Forest models are slower than Gradient Boosted Trees models. However, YDF Random Forest inference is slighly slower than scikit-learn.
    \item YDF with ``benchmark hp'', which notably includes oblique splits, are significantly slower to train and to infer than the ``default'' version.
    \item The algorithmic complexity of inference of the linear model is significantly less than that of the decision forest models. Surprisingly, the inference of linear models executed through TensorFlow are slowest.
\end{itemize}

\section{Conclusion}
\label{sec:conclusion}

We presented a new library for the training, inference, and interpretation of decision forest models called Yggdrasil decision forests. The library is designed around four principles to ensure extensibility to new methods and efficient usage. While designed for this library, we believe these principles carry over to other machine learning libraries. We showed how to use the CLI and Python APIs of Yggdrasil and empirically compared its quality and speed with existing libraries.

\newpage

\bibliography{lib}


\begin{thebibliography}{32}


\ifx \showCODEN    \undefined \def \showCODEN     #1{\unskip}     \fi
\ifx \showDOI      \undefined \def \showDOI       #1{#1}\fi
\ifx \showISBNx    \undefined \def \showISBNx     #1{\unskip}     \fi
\ifx \showISBNxiii \undefined \def \showISBNxiii  #1{\unskip}     \fi
\ifx \showISSN     \undefined \def \showISSN      #1{\unskip}     \fi
\ifx \showLCCN     \undefined \def \showLCCN      #1{\unskip}     \fi
\ifx \shownote     \undefined \def \shownote      #1{#1}          \fi
\ifx \showarticletitle \undefined \def \showarticletitle #1{#1}   \fi
\ifx \showURL      \undefined \def \showURL       {\relax}        \fi
\providecommand\bibfield[2]{#2}
\providecommand\bibinfo[2]{#2}
\providecommand\natexlab[1]{#1}
\providecommand\showeprint[2][]{arXiv:#2}

\bibitem[Abadi et~al\mbox{.}(2016)]%
        {tensorflow_2015}
\bibfield{author}{\bibinfo{person}{Mart{\'\i}n Abadi}, \bibinfo{person}{Paul
  Barham}, \bibinfo{person}{Jianmin Chen}, \bibinfo{person}{Zhifeng Chen},
  \bibinfo{person}{Andy Davis}, \bibinfo{person}{Jeffrey Dean},
  \bibinfo{person}{Matthieu Devin}, \bibinfo{person}{Sanjay Ghemawat},
  \bibinfo{person}{Geoffrey Irving}, \bibinfo{person}{Michael Isard},
  {et~al\mbox{.}}} \bibinfo{year}{2016}\natexlab{}.
\newblock \showarticletitle{Tensorflow: A system for large-scale machine
  learning}. In \bibinfo{booktitle}{\emph{12th $\{$USENIX$\}$ Symposium on
  Operating Systems Design and Implementation ($\{$OSDI$\}$ 16)}}.
  \bibinfo{pages}{265--283}.
\newblock


\bibitem[Asadi et~al\mbox{.}(2014)]%
        {nima_2014}
\bibfield{author}{\bibinfo{person}{Nima Asadi}, \bibinfo{person}{Jimmy Lin},
  {and} \bibinfo{person}{Arjen~P. de Vries}.} \bibinfo{year}{2014}\natexlab{}.
\newblock \showarticletitle{Runtime Optimizations for Tree-Based Machine
  Learning Models}.
\newblock \bibinfo{journal}{\emph{IEEE Transactions on Knowledge and Data
  Engineering}} \bibinfo{volume}{26}, \bibinfo{number}{9}
  (\bibinfo{year}{2014}), \bibinfo{pages}{2281--2292}.
\newblock


\bibitem[Breiman(2001)]%
        {breiman_2001}
\bibfield{author}{\bibinfo{person}{L Breiman}.}
  \bibinfo{year}{2001}\natexlab{}.
\newblock \showarticletitle{Random Forests}.
\newblock \bibinfo{journal}{\emph{Machine Learning}}  \bibinfo{volume}{45}
  (\bibinfo{date}{10} \bibinfo{year}{2001}), \bibinfo{pages}{5--32}.
\newblock


\bibitem[Breiman et~al\mbox{.}(1984)]%
        {breiman_1984}
\bibfield{author}{\bibinfo{person}{Leo Breiman}, \bibinfo{person}{Jerome~H
  Friedman}, \bibinfo{person}{Richard~A Olshen}, {and}
  \bibinfo{person}{Charles~J Stone}.} \bibinfo{year}{1984}\natexlab{}.
\newblock \bibinfo{booktitle}{\emph{Classification and regression trees}}.
\newblock \bibinfo{publisher}{Routledge}.
\newblock


\bibitem[Bruch et~al\mbox{.}(2020)]%
        {bruch_2020_differentiable}
\bibfield{author}{\bibinfo{person}{Sebastian Bruch}, \bibinfo{person}{Jan
  Pfeifer}, {and} \bibinfo{person}{Mathieu Guillame-bert}.}
  \bibinfo{year}{2020}\natexlab{}.
\newblock \bibinfo{title}{Learning Representations for Axis-Aligned Decision
  Forests through Input Perturbation}.
\newblock
\newblock


\bibitem[Buitinck et~al\mbox{.}(2013)]%
        {sklearn_api}
\bibfield{author}{\bibinfo{person}{Lars Buitinck}, \bibinfo{person}{Gilles
  Louppe}, \bibinfo{person}{Mathieu Blondel}, \bibinfo{person}{Fabian
  Pedregosa}, \bibinfo{person}{Andreas Mueller}, \bibinfo{person}{Olivier
  Grisel}, \bibinfo{person}{Vlad Niculae}, \bibinfo{person}{Peter
  Prettenhofer}, \bibinfo{person}{Alexandre Gramfort}, \bibinfo{person}{Jaques
  Grobler}, \bibinfo{person}{Robert Layton}, \bibinfo{person}{Jake VanderPlas},
  \bibinfo{person}{Arnaud Joly}, \bibinfo{person}{Brian Holt}, {and}
  \bibinfo{person}{Ga{\"{e}}l Varoquaux}.} \bibinfo{year}{2013}\natexlab{}.
\newblock \showarticletitle{{API} design for machine learning software:
  experiences from the scikit-learn project}. In \bibinfo{booktitle}{\emph{ECML
  PKDD Workshop: Languages for Data Mining and Machine Learning}}.
  \bibinfo{pages}{108--122}.
\newblock


\bibitem[Chen and Guestrin(2016)]%
        {chen_2016}
\bibfield{author}{\bibinfo{person}{Tianqi Chen} {and} \bibinfo{person}{Carlos
  Guestrin}.} \bibinfo{year}{2016}\natexlab{}.
\newblock \showarticletitle{{XGBoost}: A Scalable Tree Boosting System}. In
  \bibinfo{booktitle}{\emph{Proceedings of the 22nd ACM SIGKDD International
  Conference on Knowledge Discovery and Data Mining}} (San Francisco,
  California, USA) \emph{(\bibinfo{series}{KDD '16})}.
  \bibinfo{publisher}{ACM}, \bibinfo{address}{New York, NY, USA},
  \bibinfo{pages}{785--794}.
\newblock
\showISBNx{978-1-4503-4232-2}


\bibitem[Fisher(1958)]%
        {fisher_1958_exact}
\bibfield{author}{\bibinfo{person}{Walter~D. Fisher}.}
  \bibinfo{year}{1958}\natexlab{}.
\newblock \showarticletitle{On Grouping for Maximum Homogeneity}.
\newblock \bibinfo{journal}{\emph{J. Amer. Statist. Assoc.}}
  \bibinfo{volume}{53}, \bibinfo{number}{284} (\bibinfo{year}{1958}),
  \bibinfo{pages}{789--798}.
\newblock


\bibitem[Friedman(2001)]%
        {friedman_2001}
\bibfield{author}{\bibinfo{person}{Jerome~H. Friedman}.}
  \bibinfo{year}{2001}\natexlab{}.
\newblock \showarticletitle{{Greedy function approximation: A gradient boosting
  machine.}}
\newblock \bibinfo{journal}{\emph{The Annals of Statistics}}
  \bibinfo{volume}{29}, \bibinfo{number}{5} (\bibinfo{year}{2001}),
  \bibinfo{pages}{1189 -- 1232}.
\newblock


\bibitem[Guillame-Bert et~al\mbox{.}(2020)]%
        {guillame_bert_2020_catset}
\bibfield{author}{\bibinfo{person}{Mathieu Guillame-Bert},
  \bibinfo{person}{Sebastian Bruch}, \bibinfo{person}{Petr Mitrichev},
  \bibinfo{person}{Petr Mikheev}, {and} \bibinfo{person}{Jan Pfeifer}.}
  \bibinfo{year}{2020}\natexlab{}.
\newblock \bibinfo{title}{Modeling Text with Decision Forests using
  Categorical-Set Splits}.
\newblock
\newblock
\urldef\tempurl%
\url{https://doi.org/10.48550/ARXIV.2009.09991}
\showDOI{\tempurl}


\bibitem[Guillame{-}Bert and Teytaud(2018)]%
        {guillame_bert_2018}
\bibfield{author}{\bibinfo{person}{Mathieu Guillame{-}Bert} {and}
  \bibinfo{person}{Olivier Teytaud}.} \bibinfo{year}{2018}\natexlab{}.
\newblock \showarticletitle{Exact Distributed Training: Random Forest with
  Billions of Examples}.
\newblock \bibinfo{journal}{\emph{CoRR}}  \bibinfo{volume}{abs/1804.06755}
  (\bibinfo{year}{2018}).
\newblock
\showeprint[arXiv]{1804.06755}


\bibitem[Harris et~al\mbox{.}(2020)]%
        {harris_2020}
\bibfield{author}{\bibinfo{person}{Charles~R. Harris},
  \bibinfo{person}{K.~Jarrod Millman}, \bibinfo{person}{Stéfan~J van~der
  Walt}, \bibinfo{person}{Ralf Gommers}, \bibinfo{person}{Pauli Virtanen},
  \bibinfo{person}{David Cournapeau}, \bibinfo{person}{Eric Wieser},
  \bibinfo{person}{Julian Taylor}, \bibinfo{person}{Sebastian Berg},
  \bibinfo{person}{Nathaniel~J. Smith}, \bibinfo{person}{Robert Kern},
  \bibinfo{person}{Matti Picus}, \bibinfo{person}{Stephan Hoyer},
  \bibinfo{person}{Marten~H. van Kerkwijk}, \bibinfo{person}{Matthew Brett},
  \bibinfo{person}{Allan Haldane}, \bibinfo{person}{Jaime Fernández~del Río},
  \bibinfo{person}{Mark Wiebe}, \bibinfo{person}{Pearu Peterson},
  \bibinfo{person}{Pierre Gérard-Marchant}, \bibinfo{person}{Kevin Sheppard},
  \bibinfo{person}{Tyler Reddy}, \bibinfo{person}{Warren Weckesser},
  \bibinfo{person}{Hameer Abbasi}, \bibinfo{person}{Christoph Gohlke}, {and}
  \bibinfo{person}{Travis~E. Oliphant}.} \bibinfo{year}{2020}\natexlab{}.
\newblock \showarticletitle{Array programming with {NumPy}}.
\newblock \bibinfo{journal}{\emph{Nature}}  \bibinfo{volume}{585}
  (\bibinfo{year}{2020}), \bibinfo{pages}{357–362}.
\newblock


\bibitem[Ke et~al\mbox{.}(2017)]%
        {ke_2017_lightgbm}
\bibfield{author}{\bibinfo{person}{Guolin Ke}, \bibinfo{person}{Qi Meng},
  \bibinfo{person}{Thomas Finley}, \bibinfo{person}{Taifeng Wang},
  \bibinfo{person}{Wei Chen}, \bibinfo{person}{Weidong Ma},
  \bibinfo{person}{Qiwei Ye}, {and} \bibinfo{person}{Tie-Yan Liu}.}
  \bibinfo{year}{2017}\natexlab{}.
\newblock \showarticletitle{Lightgbm: A highly efficient gradient boosting
  decision tree}.
\newblock \bibinfo{journal}{\emph{Advances in neural information processing
  systems}}  \bibinfo{volume}{30} (\bibinfo{year}{2017}),
  \bibinfo{pages}{3146--3154}.
\newblock


\bibitem[Ke et~al\mbox{.}(2019)]%
        {guolin_2019_deep_gbm}
\bibfield{author}{\bibinfo{person}{Guolin Ke}, \bibinfo{person}{Zhenhui Xu},
  \bibinfo{person}{Jia Zhang}, \bibinfo{person}{Jiang Bian}, {and}
  \bibinfo{person}{Tie-Yan Liu}.} \bibinfo{year}{2019}\natexlab{}.
\newblock \showarticletitle{DeepGBM: A Deep Learning Framework Distilled by
  GBDT for Online Prediction Tasks}. In \bibinfo{booktitle}{\emph{Proceedings
  of the 25th ACM SIGKDD International Conference on Knowledge Discovery \&
  Data Mining}} (Anchorage, AK, USA) \emph{(\bibinfo{series}{KDD '19})}.
  \bibinfo{publisher}{Association for Computing Machinery},
  \bibinfo{pages}{384–394}.
\newblock


\bibitem[Leo~Breiman(1984)]%
        {breiman_1984_cart}
\bibfield{author}{\bibinfo{person}{Charles J. Stone R.A.~Olshen Leo~Breiman,
  Jerome~Friedman}.} \bibinfo{year}{1984}\natexlab{}.
\newblock \bibinfo{booktitle}{\emph{Classification and Regression Trees}}.
\newblock \bibinfo{publisher}{Chapman and Hall/CRC}.
\newblock


\bibitem[Li et~al\mbox{.}(2019)]%
        {li_2019_combine}
\bibfield{author}{\bibinfo{person}{Pan Li}, \bibinfo{person}{Zhen Qin},
  \bibinfo{person}{Xuanhui Wang}, {and} \bibinfo{person}{Don Metzler}.}
  \bibinfo{year}{2019}\natexlab{}.
\newblock \showarticletitle{Combining Decision Trees and Neural Networks for
  Learning-to-Rank in Personal Search}.
\newblock


\bibitem[Liaw and Wiener(2002)]%
        {wiener_2002}
\bibfield{author}{\bibinfo{person}{Andy Liaw} {and} \bibinfo{person}{Matthew
  Wiener}.} \bibinfo{year}{2002}\natexlab{}.
\newblock \showarticletitle{Classification and Regression by randomForest}.
\newblock \bibinfo{journal}{\emph{R News}} \bibinfo{volume}{2},
  \bibinfo{number}{3} (\bibinfo{year}{2002}), \bibinfo{pages}{18--22}.
\newblock


\bibitem[Lucchese et~al\mbox{.}(2015)]%
        {lucchese_2015}
\bibfield{author}{\bibinfo{person}{Claudio Lucchese},
  \bibinfo{person}{Franco~Maria Nardini}, \bibinfo{person}{Salvatore Orlando},
  \bibinfo{person}{Raffaele Perego}, \bibinfo{person}{Nicola Tonellotto}, {and}
  \bibinfo{person}{Rossano Venturini}.} \bibinfo{year}{2015}\natexlab{}.
\newblock \showarticletitle{QuickScorer: A Fast Algorithm to Rank Documents
  with Additive Ensembles of Regression Trees}. In
  \bibinfo{booktitle}{\emph{Proceedings of the 38th International ACM SIGIR
  Conference on Research and Development in Information Retrieval}} (Santiago,
  Chile) \emph{(\bibinfo{series}{SIGIR '15})}. \bibinfo{publisher}{Association
  for Computing Machinery}, \bibinfo{address}{New York, NY, USA},
  \bibinfo{pages}{73–82}.
\newblock


\bibitem[Lütkebohle(2015)]%
        {bazel_2015}
\bibfield{author}{\bibinfo{person}{Ingo Lütkebohle}.}
  \bibinfo{year}{2015}\natexlab{}.
\newblock \bibinfo{title}{Bazel build tool}.
\newblock \bibinfo{howpublished}{\url{https://bazel.build}}.
\newblock
\newblock
\shownote{[Online; accessed 12-September-2022]}.


\bibitem[Mathieu Guillame-Bert({[n.\,d.]})]%
        {sml_for_sheets}
\bibfield{author}{\bibinfo{person}{Richard Stotz Luiz GUStavo Martins Ashley
  Oldacre Jocelyn Becker Glenn~Cameron Mathieu Guillame-Bert, Jan~Pfeifer}.}
  \bibinfo{year}{[n.\,d.]}\natexlab{}.
\newblock \showarticletitle{Simple ML for Sheets}.
\newblock  (\bibinfo{year}{[n.\,d.]}).
\newblock
\urldef\tempurl%
\url{https://simplemlforsheets.com/}
\showURL{%
\tempurl}


\bibitem[Ponomareva et~al\mbox{.}(2017)]%
        {ponomareva_2017_tfb}
\bibfield{author}{\bibinfo{person}{Natalia Ponomareva},
  \bibinfo{person}{Soroush Radpour}, \bibinfo{person}{Gilbert Hendry},
  \bibinfo{person}{Salem Haykal}, \bibinfo{person}{Thomas Colthurst},
  \bibinfo{person}{Petr Mitrichev}, {and} \bibinfo{person}{Alexander
  Grushetsky}.} \bibinfo{year}{2017}\natexlab{}.
\newblock \showarticletitle{Tf boosted trees: A scalable tensorflow based
  framework for gradient boosting}. In \bibinfo{booktitle}{\emph{Joint European
  Conference on Machine Learning and Knowledge Discovery in Databases}}.
  Springer, \bibinfo{pages}{423--427}.
\newblock


\bibitem[Popov et~al\mbox{.}(2019)]%
        {popov_2019_xgboost_experiment}
\bibfield{author}{\bibinfo{person}{Sergei Popov}, \bibinfo{person}{Stanislav
  Morozov}, {and} \bibinfo{person}{Artem Babenko}.}
  \bibinfo{year}{2019}\natexlab{}.
\newblock \showarticletitle{Neural oblivious decision ensembles for deep
  learning on tabular data}.
\newblock \bibinfo{journal}{\emph{arXiv preprint arXiv:1909.06312}}
  (\bibinfo{year}{2019}).
\newblock


\bibitem[Quinlan(1994)]%
        {quinlan_1994}
\bibfield{author}{\bibinfo{person}{John Quinlan}.}
  \bibinfo{year}{1994}\natexlab{}.
\newblock \bibinfo{title}{C4. 5: Programs for machine learning}.
\newblock
\newblock


\bibitem[{R Core Team}(2022)]%
        {R_2022}
\bibfield{author}{\bibinfo{person}{{R Core Team}}.}
  \bibinfo{year}{2022}\natexlab{}.
\newblock \bibinfo{booktitle}{\emph{R: A Language and Environment for
  Statistical Computing}}.
\newblock R Foundation for Statistical Computing, Vienna, Austria.
\newblock
\urldef\tempurl%
\url{https://www.R-project.org/}
\showURL{%
\tempurl}


\bibitem[Schapire(2013)]%
        {schapire_2013}
\bibfield{author}{\bibinfo{person}{Robert~E Schapire}.}
  \bibinfo{year}{2013}\natexlab{}.
\newblock \showarticletitle{Explaining adaboost}.
\newblock In \bibinfo{booktitle}{\emph{Empirical inference}}.
  \bibinfo{publisher}{Springer}, \bibinfo{pages}{37--52}.
\newblock


\bibitem[Shi(2007)]%
        {shi_2007}
\bibfield{author}{\bibinfo{person}{Haijia Shi}.}
  \bibinfo{year}{2007}\natexlab{}.
\newblock \showarticletitle{Best-first Decision Tree Learning}.
\newblock


\bibitem[Sullivan et~al\mbox{.}(2001)]%
        {sullivan_2001}
\bibfield{author}{\bibinfo{person}{Kevin~J. Sullivan},
  \bibinfo{person}{William~G. Griswold}, \bibinfo{person}{Yuanfang Cai}, {and}
  \bibinfo{person}{Ben Hallen}.} \bibinfo{year}{2001}\natexlab{}.
\newblock \showarticletitle{The Structure and Value of Modularity in Software
  Design}.
\newblock \bibinfo{journal}{\emph{SIGSOFT Softw. Eng. Notes}}
  \bibinfo{volume}{26}, \bibinfo{number}{5} (\bibinfo{date}{sep}
  \bibinfo{year}{2001}), \bibinfo{pages}{99–108}.
\newblock


\bibitem[Terry~Therneau and Ripley({[n.\,d.]})]%
        {therneau_2002}
\bibfield{author}{\bibinfo{person}{Beth~Atkinson Terry~Therneau} {and}
  \bibinfo{person}{Brian Ripley}.} \bibinfo{year}{[n.\,d.]}\natexlab{}.
\newblock \showarticletitle{rpart: Recursive Partitioning and Regression
  Trees}.
\newblock  (\bibinfo{year}{[n.\,d.]}).
\newblock
\urldef\tempurl%
\url{https://CRAN.R-project.org/package=rpart}
\showURL{%
\tempurl}


\bibitem[Tomita et~al\mbox{.}(2020)]%
        {tomita_2020}
\bibfield{author}{\bibinfo{person}{Tyler~M. Tomita}, \bibinfo{person}{James
  Browne}, \bibinfo{person}{Cencheng Shen}, \bibinfo{person}{Jaewon Chung},
  \bibinfo{person}{Jesse~L. Patsolic}, \bibinfo{person}{Benjamin Falk},
  \bibinfo{person}{Carey~E. Priebe}, \bibinfo{person}{Jason Yim},
  \bibinfo{person}{Randal Burns}, \bibinfo{person}{Mauro Maggioni}, {and}
  \bibinfo{person}{Joshua~T. Vogelstein}.} \bibinfo{year}{2020}\natexlab{}.
\newblock \showarticletitle{Sparse Projection Oblique Randomer Forests}.
\newblock \bibinfo{journal}{\emph{J. Mach. Learn. Res.}} \bibinfo{volume}{21},
  \bibinfo{number}{1}, Article \bibinfo{articleno}{104} (\bibinfo{date}{jan}
  \bibinfo{year}{2020}), \bibinfo{numpages}{39}~pages.
\newblock


\bibitem[Vanschoren et~al\mbox{.}(2014)]%
        {vanschoren_2013_open_ml}
\bibfield{author}{\bibinfo{person}{Joaquin Vanschoren}, \bibinfo{person}{Jan~N.
  van Rijn}, \bibinfo{person}{Bernd Bischl}, {and} \bibinfo{person}{Luis
  Torgo}.} \bibinfo{year}{2014}\natexlab{}.
\newblock \showarticletitle{OpenML: Networked Science in Machine Learning}.
\newblock \bibinfo{journal}{\emph{SIGKDD Explor. Newsl.}} \bibinfo{volume}{15},
  \bibinfo{number}{2} (\bibinfo{date}{jun} \bibinfo{year}{2014}),
  \bibinfo{pages}{49–60}.
\newblock


\bibitem[{W}es {M}c{K}inney(2010)]%
        {mckinney_2010}
\bibfield{author}{\bibinfo{person}{{W}es {M}c{K}inney}.}
  \bibinfo{year}{2010}\natexlab{}.
\newblock \showarticletitle{{D}ata {S}tructures for {S}tatistical {C}omputing
  in {P}ython}. In \bibinfo{booktitle}{\emph{{P}roceedings of the 9th {P}ython
  in {S}cience {C}onference}}, \bibfield{editor}{\bibinfo{person}{{S}t\'efan
  van~der {W}alt} {and} \bibinfo{person}{{J}arrod {M}illman}} (Eds.).
  \bibinfo{pages}{56 -- 61}.
\newblock


\bibitem[Wright and Ziegler(2017)]%
        {wright_2017}
\bibfield{author}{\bibinfo{person}{Marvin~N. Wright} {and}
  \bibinfo{person}{Andreas Ziegler}.} \bibinfo{year}{2017}\natexlab{}.
\newblock \showarticletitle{ranger: A Fast Implementation of Random Forests for
  High Dimensional Data in {C++} and {R}}.
\newblock \bibinfo{journal}{\emph{Journal of Statistical Software}}
  \bibinfo{volume}{77}, \bibinfo{number}{1} (\bibinfo{year}{2017}),
  \bibinfo{pages}{1--17}.
\newblock


\end{thebibliography}
\bibliographystyle{ACM-Reference-Format}

\appendix

\section{Naïve decision tree inference}
\label{alg:default_serving}

\begin{algorithm}[!h]
   \caption{Simple tree inference algorithm}
\begin{algorithmic}
   \STATE {\bfseries Input:} example $x$
   \STATE $c \leftarrow $ root node of the tree
   \WHILE{$c$ is not a leaf}{}
   \STATE $e \leftarrow $ evaluate the condition of $c$ on $x$
   \IF{$e$ \bf{is} \textsc{True}}
   \STATE $c \leftarrow $ \textsc{Positive Child of} $c$
   \ELSE
   \STATE $c \leftarrow $ \textsc{Negative Child of} $c$
   \ENDIF
   \ENDWHILE
   \STATE \textbf{return} $c$'s value
\end{algorithmic}
\end{algorithm}

\section{Artefacts from the CLI usage example}
\label{sec:artefacts_cli}

\subsection{Column information}
\label{sec:artefacts_cli_dataspec}

Sample from the column information (dataspec) returned by \emph{show\_dataspec}.

\begin{small}
\begin{spverbatim}
Number of records: 22792
Number of columns: 15

Number of columns by type:
    CATEGORICAL: 9 (60%)
    NUMERICAL: 6 (40%)

Columns:

CATEGORICAL: 9 (60%)
    3: "education" CATEGORICAL has-dict vocab-size:17 zero-ood-items most-frequent:"HS-grad" 7340 (32.2043%)
    14: "income" CATEGORICAL has-dict vocab-size:3 zero-ood-items most-frequent:"<=50K" 17308 (75.9389%)
    5: "marital_status" CATEGORICAL has-dict vocab-size:8 zero-ood-items most-frequent:"Married-civ-spouse" 10431 (45.7661%)
    ...

NUMERICAL: 6 (40%)
    0: "age" NUMERICAL mean:38.6153 min:17 max:90 sd:13.661
    10: "capital_gain" NUMERICAL mean:1081.9 min:0 max:99999 sd:7509.48
    11: "capital_loss" NUMERICAL mean:87.2806 min:0 max:4356 sd:403.01
    ...

Terminology:
    nas: Number of non-available (i.e. missing) values.
    ood: Out of dictionary.
    manually-defined: Attribute which type is manually defined by the user i.e. the type was not automatically inferred.
    tokenized: The attribute value is obtained through tokenization.
    has-dict: The attribute is attached to a string dictionary e.g. a categorical attribute stored as a string.
    vocab-size: Number of unique values.
\end{spverbatim}
\end{small}

\subsection{Model information}
\label{sec:artefacts_cli_info}

Sample from the result returned by \emph{show\_model}.

\begin{small}
\begin{spverbatim}
Type: "GRADIENT_BOOSTED_TREES"
Task: CLASSIFICATION
Label: "income"

Input Features (14):
    age
    workclass
    fnlwgt
    ...
    hours_per_week
    native_country

Variable Importance: NUM_AS_ROOT:
    1.            "age" 37 ###############
    2.   "capital_gain" 24 #########
    3. "marital_status" 22 ########
    4. "hours_per_week" 21 #######
    ...

Variable Importance: NUM_NODES:
    1.     "occupation" 703 ###############
    2.         "fnlwgt" 659 #############
    3.      "education" 614 ############
    4.            "age" 557 ###########
    ...

Loss: BINOMIAL_LOG_LIKELIHOOD
Validation loss value: 0.578763
Number of trees per iteration: 1
Number of trees: 186
Total number of nodes: 9898

Number of nodes by tree:
Count: 186 Average: 53.2151 StdDev: 7.25844
Min: 23 Max: 63 Ignored: 0
----------------------------------------------
[ 23, 25)  1   0.54%   0.54%
[ 25, 27)  0   0.00%   0.54%
[ 27, 29)  0   0.00%   0.54%
...
[ 57, 59) 25  13.44%  70.97% ########
[ 59, 61) 23  12.37%  83.33% #######
[ 61, 63] 31  16.67% 100.00% ##########

Depth by leafs:
Count: 5042 Average: 4.86573 StdDev: 0.443584
Min: 2 Max: 5 Ignored: 0
----------------------------------------------
[ 2, 3)   30   0.60%   0.60%
[ 3, 4)  113   2.24%   2.84%
[ 4, 5)  361   7.16%  10.00% #
[ 5, 5] 4538  90.00% 100.00% ##########

Number of training obs by leaf:
Count: 5042 Average: 757.465 StdDev: 2542.45
Min: 5 Max: 20094 Ignored: 0
----------------------------------------------
[     5,  1009) 4426  87.78%  87.78% ####
[  1009,  2014)  192   3.81%  91.59%
...
[  8041,  9045)   11   0.22%  97.42%
[  9045, 10050)   11   0.22%  97.64%

Attribute in nodes:
    703 : occupation [CATEGORICAL]
    659 : fnlwgt [NUMERICAL]
    614 : education [CATEGORICAL]
    ...
    47 : sex [CATEGORICAL]
    34 : race [CATEGORICAL]

Attribute in nodes with depth <= 0:
    37 : age [NUMERICAL]
    24 : capital_gain [NUMERICAL]
    22 : marital_status [CATEGORICAL]
    ...
    3 : workclass [CATEGORICAL]
    2 : occupation [CATEGORICAL]

Attribute in nodes with depth <= 1:
    80 : fnlwgt [NUMERICAL]
    77 : capital_gain [NUMERICAL]
    73 : age [NUMERICAL]
    ...
    33 : hours_per_week [NUMERICAL]
    29 : relationship [CATEGORICAL]

Condition type in nodes:
        2509 : ContainsBitmapCondition
        2342 : HigherCondition
        5 : ContainsCondition
...
\end{spverbatim}
\end{small}

\subsection{Model evaluation report}
\label{sec:artefacts_cli_eval}

\begin{figure}[t]
\centering
\includegraphics[scale=0.4]{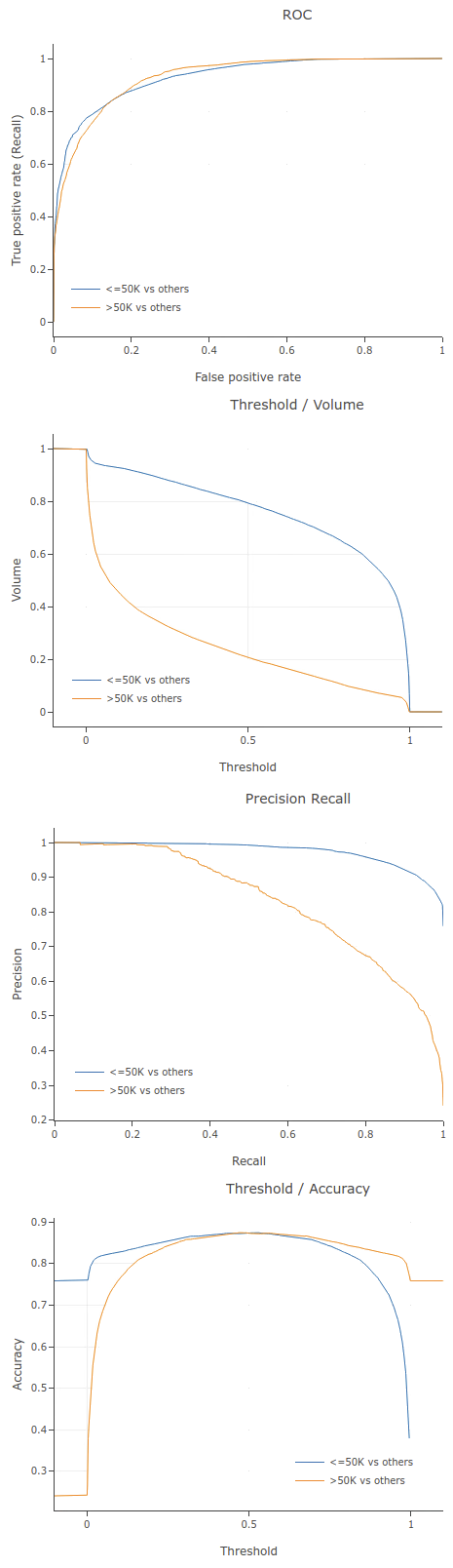}
\caption{Example of model evaluation report plots}\label{fig:evaluation_report}
\end{figure}

The evaluation report is composed of both text and plots (fig.\ref{fig:evaluation_report}). The dataset was evaluated on 9769 examples for an accurate of 0.8734, and with a 95\% confidence boundaries of [0.8678 0.8789] computed with bootstrapping (W). Other common binary classification metrics are reported (AUC, PR-AUC, AP, logloss, error rate).

\begin{small}
\begin{verbatim}
Evaluation:                                                                        
Number of predictions (without weights): 9769                                      
Number of predictions (with weights): 9769                                         
Task: CLASSIFICATION                                                               
Label: income                                                                      

Accuracy: 0.873477  CI95[W][0.867811 0.878978]                                     
LogLoss: 0.277841                                                                  
ErrorRate: 0.126523                                                                

Default Accuracy: 0.758727                                                         
Default LogLoss: 0.552543                                                          
Default ErrorRate: 0.241273                                                        

Confusion Table:                                                                   
truth\prediction                                                                   
       <OOD>  <=50K  >50K 
<OOD>      0      0     0                                                          
<=50K      0   6962   450 
 >50K      0    786  1571                                                          
Total: 9769                                                                        

One vs other classes:                                                              
  "<=50K" vs. the others                                                           
    auc: 0.929051  CI95[H][0.92419 0.93390]
                   CI95[B][0.92355 0.93420]           
    p/r-auc: 0.9756 CI95[L][0.97184 0.97888]
                    CI95[B][0.97323 0.97791]        
    ap: 0.975609   CI95[B][0.97323 0.97790]

  ">50K" vs. the others
    auc: 0.929051  CI95[H][0.92170 0.9364]
                   CI95[B][0.92369 0.93454]
    p/r-auc: 0.83002 CI95[L][0.81431 0.8446]
                     CI95[B][0.81766 0.8431]
    ap: 0.829984   CI95[B][0.8175 0.84308]
\end{verbatim}
\end{small}

\subsection{Model inference benchmark report}

Following is the inference benchmark report. The model was run 20 times over the whole dataset. Three engines have been found compatible with the model. The fastest one, called GradientBoostedTreesQuickScorer shows an average inference time of $1.066 \mu s$ per examples on a single CPU:

\begin{small}
\begin{verbatim}
batch_size : 100  num_runs : 20
time/example(us)  time/batch(us)  method
----------------------------------------
1.066 106.6 GradientBoostedTreesQuickScorer
6.7385 673.85 GradientBoostedTreesGeneric
15.829 1582.9 Generic slow engine
---------------------------------------- 
\end{verbatim}
\end{small}

\section{Experiments}

\subsection{Default and \texttt{benchmark\_rank1@v1} hyper-parameters}
\label{sec:default_hyper_paramters}

The following hyper-parameters have been used for non-hyper-parameter tuned learners.

\textbf{Random Forest default hyper-parameters}

\begin{itemize}
\item categorical\_algorithm: CART
\item growing\_strategy: LOCAL
\item max\_depth: 16
\item min\_examples: 5
\item num\_candidate\_attributes: Breiman rule of thumb
\item split\_axis: AXIS\_ALIGNED
\end{itemize}

\textbf{Random Forest rank1@v1 hyper-parameters}

Same as the default hyper-parameters with the following changes.

\begin{itemize}
\item categorical\_algorithm: RANDOM
\item split\_axis: SPARSE\_OBLIQUE
\item sparse\_oblique\_normalization: MIN\_MAX
\item sparse\_oblique\_num\_projections\_exponent: 1
\end{itemize}

\textbf{Gradient Boosted Trees hyper-parameters}

\begin{itemize}
\item early\_stopping: LOSS\_INCREASE
\item l1\_regularization: 0
\item l2\_regularization: 0
\item max\_depth: 6
\item num\_candidate\_attributes: -1 i.e. all
\item shrinkage: 0.1
\item sampling\_method: NONE
\item use\_hessian\_gain: false
\item growing\_strategy: LOCAL
\item categorical\_algorithm: CART
\item split\_axis: AXIS\_ALIGNED
\end{itemize}

\textbf{Gradient Boosted rank1@v1 hyper-parameters}

Same as the default hyper-parameters with the following changes.

\begin{itemize}
\item growing\_strategy: BEST\_FIRST\_GLOBAL
\item categorical\_algorithm: RANDOM
\item split\_axis: SPARSE\_OBLIQUE
\item sparse\_oblique\_normalization: MIN\_MAX
\item sparse\_oblique\_num\_projections\_exponent: 1
\end{itemize}

\subsection{Hyper-parameter space for hyper-parameter tuning}
\label{sec:benchmark_hyper_paramters}

The following hyper-parameter were considered during automatic hyper-parameter optimization. 

\begin{description}
\item[Yggdrasil Decision Forests]:  Min sample per leaf: 2 to 10. Categorical splitting algorithm: CART or Random. Splits: Axis aligned or Sparse. Random Forest: Max depth: 12 to 30. Gradient Boosted Trees: Hessian splits: Yes or No. Shrinkage: 0.02 to 0.15. Num candidate attribute ratio: 0.2 to 1.0. Growing strategy: divide-and-conquer (with max depth in 3 to 8) or global (with max num nodes in 16 to 256).
\item[LightGBM]: We use similar hyper-parameters as for YDF as well as remarks regarding parameter tuning from the LightGBM documentation. Gradient Boosted Trees: Num leafs: 16 to 256. Min data in leaf: 2 to 10. Max depth: 3 to 8. Learning rate: 0.02 to 0.15. Bagging fraction: 0.5 to 1.0. Feature faction: 0.2 to 1.0.
\item[XGBoost]: We use similar hyper-parameters as used in~\cite{popov_2019_xgboost_experiment} as well as remarks regarding parameter tuning on the XGBoost documentation.  Gradient Boosted Trees: Eta (shrinkage): 0.002 to 0.015. Max depth: 2 to 9. Subsample: 0.5 to 1.0. ColSample by tree: 0.2 t o 1.0. Min child weight: 2 to 10.
\item[Scikit-learn]: Random Forest: Max depth: 12 to 30. Min sample per leaf: 1 to 40.
\end{description}

\subsection{Pair wise learner comparison}
\label{sec:benchmark_pairwise}

Tab.~\ref{tab:pair_wise_acc} shows the pairwise comparison between each learners regarding accuracy.

\subsection{Accuracies}
\label{sec:benchmark_accuracy}

Tab.~\ref{tab:accuracies} shows the accuracy for each learner on each dataset. The reported values are the average over the 10 cross-validation rounds. Learners are sorted by average global rank (the first learners are the better).

\subsection{Training time}
\label{sec:benchmark_training_time}

Tab.~\ref{tab:training_time} shows the training time of each learner on each dataset. The reported values are the average over the 10 cross-validation rounds.

\subsection{Inference time}
\label{sec:benchmark_inference_time}

Tab.~\ref{tab:inference_time} shows the inference time of each learner on each dataset. The reported values are the average over the 10 cross-validation rounds.

\subsection{Datasets}
\label{sec:benchmark_dataset_stats}

Tab.~\ref{tab:dataset_stats} shows statistics about each dataset. The reported values are the average over the 10 cross-validation rounds.

{
\onecolumn
\begin{sidewaystable}
\setlength{\tabcolsep}{2pt}
\vskip 5.0in
\caption{Pairwise comparison between each learners regarding accuracy. Each cell contains the ``number of wins / number of losses (average accuracy difference)" of the row learner compared to the column learner over all the datasets and all the cross-validation folds. The number of comparisons (sum of wins and losses) is 'number of datasets x number folds'. Ties are counted as 0.5 win / 0.5 loss. For example, a large number of wins and a large positive mean metric difference indicates that the row learner is better that the column learner. Learners are sorted by average global rank (the first learners are the better). The cell is green iif. more than half of the pairwise comparisons are won. Note that the number of wins and the cell color does not take into account the among of metric difference (only which one is better).\label{tab:pair_wise_acc}}
\small
\begin{tabular}{rl|rrrrrrrrrrrrrrrr}
\hline
 &  & 1 & 2 & 3 & 4 & 5 & 6 & 7 & 8 & 9 & 10 & 11 & 12 & 13 & 14 & 15 & 16 \\ \hline
1 & YDF Autotuned (opt loss) &  & \cellcolor[HTML]{FF4D4D}341/359 & \cellcolor[HTML]{55FF00}377/323 & \cellcolor[HTML]{55FF00}396/305 & \cellcolor[HTML]{55FF00}394/306 & \cellcolor[HTML]{55FF00}428/273 & \cellcolor[HTML]{55FF00}457/243 & \cellcolor[HTML]{55FF00}442/258 & \cellcolor[HTML]{55FF00}465/235 & \cellcolor[HTML]{55FF00}524/177 & \cellcolor[HTML]{55FF00}498/202 & \cellcolor[HTML]{55FF00}563/138 & \cellcolor[HTML]{55FF00}585/116 & \cellcolor[HTML]{55FF00}605/95 & \cellcolor[HTML]{55FF00}600/100 & \cellcolor[HTML]{55FF00}612/88 \\
2 & YDF Autotuned (opt acc) & \cellcolor[HTML]{55FF00}359/341 &  & \cellcolor[HTML]{55FF00}385/315 & \cellcolor[HTML]{55FF00}402/298 & \cellcolor[HTML]{55FF00}406/294 & \cellcolor[HTML]{55FF00}432/269 & \cellcolor[HTML]{55FF00}458/243 & \cellcolor[HTML]{55FF00}435/265 & \cellcolor[HTML]{55FF00}465/235 & \cellcolor[HTML]{55FF00}499/201 & \cellcolor[HTML]{55FF00}508/192 & \cellcolor[HTML]{55FF00}554/147 & \cellcolor[HTML]{55FF00}580/120 & \cellcolor[HTML]{55FF00}596/105 & \cellcolor[HTML]{55FF00}594/107 & \cellcolor[HTML]{55FF00}602/98 \\
3 & LGBM Autotuned (opt loss) & \cellcolor[HTML]{FF4D4D}323/377 & \cellcolor[HTML]{FF4D4D}315/385 &  & \cellcolor[HTML]{FF4D4D}345/355 & \cellcolor[HTML]{55FF00}372/328 & \cellcolor[HTML]{55FF00}392/308 & \cellcolor[HTML]{55FF00}419/281 & \cellcolor[HTML]{55FF00}407/294 & \cellcolor[HTML]{55FF00}442/258 & \cellcolor[HTML]{55FF00}485/215 & \cellcolor[HTML]{55FF00}498/202 & \cellcolor[HTML]{55FF00}538/163 & \cellcolor[HTML]{55FF00}565/136 & \cellcolor[HTML]{55FF00}589/111 & \cellcolor[HTML]{55FF00}569/131 & \cellcolor[HTML]{55FF00}608/92 \\
4 & YDF GBT (benchmark hp) & \cellcolor[HTML]{FF4D4D}305/396 & \cellcolor[HTML]{FF4D4D}298/402 & \cellcolor[HTML]{55FF00}355/345 &  & \cellcolor[HTML]{55FF00}374/327 & \cellcolor[HTML]{55FF00}406/294 & \cellcolor[HTML]{55FF00}435/266 & \cellcolor[HTML]{55FF00}418/282 & \cellcolor[HTML]{55FF00}415/286 & \cellcolor[HTML]{55FF00}492/209 & \cellcolor[HTML]{55FF00}490/211 & \cellcolor[HTML]{55FF00}542/159 & \cellcolor[HTML]{55FF00}564/137 & \cellcolor[HTML]{55FF00}578/123 & \cellcolor[HTML]{55FF00}574/126 & \cellcolor[HTML]{55FF00}591/109 \\
5 & LGBM Autotuned (opt acc) & \cellcolor[HTML]{FF4D4D}306/394 & \cellcolor[HTML]{FF4D4D}294/406 & \cellcolor[HTML]{FF4D4D}328/372 & \cellcolor[HTML]{FF4D4D}327/374 &  & \cellcolor[HTML]{55FF00}386/315 & \cellcolor[HTML]{55FF00}418/283 & \cellcolor[HTML]{55FF00}401/300 & \cellcolor[HTML]{55FF00}435/265 & \cellcolor[HTML]{55FF00}469/231 & \cellcolor[HTML]{55FF00}485/215 & \cellcolor[HTML]{55FF00}538/163 & \cellcolor[HTML]{55FF00}563/137 & \cellcolor[HTML]{55FF00}584/116 & \cellcolor[HTML]{55FF00}560/140 & \cellcolor[HTML]{55FF00}600/100 \\
6 & SKLean RF (default hp) & \cellcolor[HTML]{FF4D4D}273/428 & \cellcolor[HTML]{FF4D4D}269/432 & \cellcolor[HTML]{FF4D4D}308/392 & \cellcolor[HTML]{FF4D4D}294/406 & \cellcolor[HTML]{FF4D4D}315/386 &  & \cellcolor[HTML]{55FF00}383/318 & \cellcolor[HTML]{FF4D4D}343/358 & \cellcolor[HTML]{55FF00}354/346 & \cellcolor[HTML]{55FF00}453/247 & \cellcolor[HTML]{55FF00}405/295 & \cellcolor[HTML]{55FF00}518/182 & \cellcolor[HTML]{55FF00}493/207 & \cellcolor[HTML]{55FF00}547/153 & \cellcolor[HTML]{55FF00}539/162 & \cellcolor[HTML]{55FF00}556/144 \\
7 & YDF RF (benchmark hp) & \cellcolor[HTML]{FF4D4D}243/457 & \cellcolor[HTML]{FF4D4D}243/458 & \cellcolor[HTML]{FF4D4D}281/419 & \cellcolor[HTML]{FF4D4D}266/435 & \cellcolor[HTML]{FF4D4D}283/418 & \cellcolor[HTML]{FF4D4D}318/383 &  & \cellcolor[HTML]{FF4D4D}332/368 & \cellcolor[HTML]{FF4D4D}347/354 & \cellcolor[HTML]{55FF00}451/249 & \cellcolor[HTML]{55FF00}421/279 & \cellcolor[HTML]{55FF00}521/180 & \cellcolor[HTML]{55FF00}503/197 & \cellcolor[HTML]{55FF00}589/112 & \cellcolor[HTML]{55FF00}588/112 & \cellcolor[HTML]{55FF00}593/107 \\
8 & SKLearn Autotuned & \cellcolor[HTML]{FF4D4D}258/442 & \cellcolor[HTML]{FF4D4D}265/435 & \cellcolor[HTML]{FF4D4D}294/407 & \cellcolor[HTML]{FF4D4D}282/418 & \cellcolor[HTML]{FF4D4D}300/401 & \cellcolor[HTML]{55FF00}358/343 & \cellcolor[HTML]{55FF00}368/332 &  & \cellcolor[HTML]{55FF00}356/344 & \cellcolor[HTML]{55FF00}466/235 & \cellcolor[HTML]{55FF00}408/293 & \cellcolor[HTML]{55FF00}525/176 & \cellcolor[HTML]{55FF00}503/198 & \cellcolor[HTML]{55FF00}570/131 & \cellcolor[HTML]{55FF00}555/145 & \cellcolor[HTML]{55FF00}575/126 \\
9 & LGBM GBT (default hp) & \cellcolor[HTML]{FF4D4D}235/465 & \cellcolor[HTML]{FF4D4D}235/465 & \cellcolor[HTML]{FF4D4D}258/442 & \cellcolor[HTML]{FF4D4D}286/415 & \cellcolor[HTML]{FF4D4D}265/435 & \cellcolor[HTML]{FF4D4D}346/354 & \cellcolor[HTML]{55FF00}354/347 & \cellcolor[HTML]{FF4D4D}344/356 &  & \cellcolor[HTML]{55FF00}426/275 & \cellcolor[HTML]{55FF00}447/254 & \cellcolor[HTML]{55FF00}505/196 & \cellcolor[HTML]{55FF00}547/154 & \cellcolor[HTML]{55FF00}561/140 & \cellcolor[HTML]{55FF00}555/145 & \cellcolor[HTML]{55FF00}582/118 \\
10 & YDF RF (default hp) & \cellcolor[HTML]{FF4D4D}177/524 & \cellcolor[HTML]{FF4D4D}201/499 & \cellcolor[HTML]{FF4D4D}215/485 & \cellcolor[HTML]{FF4D4D}209/492 & \cellcolor[HTML]{FF4D4D}231/469 & \cellcolor[HTML]{FF4D4D}247/453 & \cellcolor[HTML]{FF4D4D}249/451 & \cellcolor[HTML]{FF4D4D}235/466 & \cellcolor[HTML]{FF4D4D}275/426 &  & \cellcolor[HTML]{FF4D4D}340/360 & \cellcolor[HTML]{55FF00}495/205 & \cellcolor[HTML]{55FF00}440/261 & \cellcolor[HTML]{55FF00}540/161 & \cellcolor[HTML]{55FF00}537/163 & \cellcolor[HTML]{55FF00}554/146 \\
11 & YDF GBT (default hp) & \cellcolor[HTML]{FF4D4D}202/498 & \cellcolor[HTML]{FF4D4D}192/508 & \cellcolor[HTML]{FF4D4D}202/498 & \cellcolor[HTML]{FF4D4D}211/490 & \cellcolor[HTML]{FF4D4D}215/485 & \cellcolor[HTML]{FF4D4D}295/405 & \cellcolor[HTML]{FF4D4D}279/421 & \cellcolor[HTML]{FF4D4D}293/408 & \cellcolor[HTML]{FF4D4D}254/447 & \cellcolor[HTML]{55FF00}360/340 &  & \cellcolor[HTML]{55FF00}471/229 & \cellcolor[HTML]{55FF00}473/227 & \cellcolor[HTML]{55FF00}521/179 & \cellcolor[HTML]{55FF00}520/181 & \cellcolor[HTML]{55FF00}535/166 \\
12 & TF Linear (default hp) & \cellcolor[HTML]{FF4D4D}138/563 & \cellcolor[HTML]{FF4D4D}147/554 & \cellcolor[HTML]{FF4D4D}163/538 & \cellcolor[HTML]{FF4D4D}159/542 & \cellcolor[HTML]{FF4D4D}163/538 & \cellcolor[HTML]{FF4D4D}182/518 & \cellcolor[HTML]{FF4D4D}180/521 & \cellcolor[HTML]{FF4D4D}176/525 & \cellcolor[HTML]{FF4D4D}196/505 & \cellcolor[HTML]{FF4D4D}205/495 & \cellcolor[HTML]{FF4D4D}229/471 &  & \cellcolor[HTML]{FF4D4D}258/442 & \cellcolor[HTML]{FF4D4D}276/424 & \cellcolor[HTML]{FF4D4D}283/417 & \cellcolor[HTML]{FF4D4D}297/404 \\
13 & XGB GBT (default hp) & \cellcolor[HTML]{FF4D4D}116/585 & \cellcolor[HTML]{FF4D4D}120/580 & \cellcolor[HTML]{FF4D4D}136/565 & \cellcolor[HTML]{FF4D4D}137/564 & \cellcolor[HTML]{FF4D4D}137/563 & \cellcolor[HTML]{FF4D4D}207/493 & \cellcolor[HTML]{FF4D4D}197/503 & \cellcolor[HTML]{FF4D4D}198/503 & \cellcolor[HTML]{FF4D4D}154/547 & \cellcolor[HTML]{FF4D4D}261/440 & \cellcolor[HTML]{FF4D4D}227/473 & \cellcolor[HTML]{55FF00}442/258 &  & \cellcolor[HTML]{55FF00}449/252 & \cellcolor[HTML]{55FF00}444/257 & \cellcolor[HTML]{55FF00}461/240 \\
14 & XGB Autotuned (opt acc) & \cellcolor[HTML]{FF4D4D}95/605 & \cellcolor[HTML]{FF4D4D}105/596 & \cellcolor[HTML]{FF4D4D}111/589 & \cellcolor[HTML]{FF4D4D}123/578 & \cellcolor[HTML]{FF4D4D}116/584 & \cellcolor[HTML]{FF4D4D}153/547 & \cellcolor[HTML]{FF4D4D}112/589 & \cellcolor[HTML]{FF4D4D}131/570 & \cellcolor[HTML]{FF4D4D}140/561 & \cellcolor[HTML]{FF4D4D}161/540 & \cellcolor[HTML]{FF4D4D}179/521 & \cellcolor[HTML]{55FF00}424/276 & \cellcolor[HTML]{FF4D4D}252/449 &  & \cellcolor[HTML]{55FF00}414/286 & \cellcolor[HTML]{55FF00}376/324 \\
15 & TF EBT (default hp) & \cellcolor[HTML]{FF4D4D}100/600 & \cellcolor[HTML]{FF4D4D}107/594 & \cellcolor[HTML]{FF4D4D}131/569 & \cellcolor[HTML]{FF4D4D}126/574 & \cellcolor[HTML]{FF4D4D}140/560 & \cellcolor[HTML]{FF4D4D}162/539 & \cellcolor[HTML]{FF4D4D}112/588 & \cellcolor[HTML]{FF4D4D}145/555 & \cellcolor[HTML]{FF4D4D}145/555 & \cellcolor[HTML]{FF4D4D}163/537 & \cellcolor[HTML]{FF4D4D}181/520 & \cellcolor[HTML]{55FF00}417/283 & \cellcolor[HTML]{FF4D4D}257/444 & \cellcolor[HTML]{FF4D4D}286/414 &  & \cellcolor[HTML]{FF4D4D}305/396 \\
16 & XGB Autotuned (opt loss) & \cellcolor[HTML]{FF4D4D}88/612 & \cellcolor[HTML]{FF4D4D}98/602 & \cellcolor[HTML]{FF4D4D}92/608 & \cellcolor[HTML]{FF4D4D}109/591 & \cellcolor[HTML]{FF4D4D}100/600 & \cellcolor[HTML]{FF4D4D}144/556 & \cellcolor[HTML]{FF4D4D}107/593 & \cellcolor[HTML]{FF4D4D}126/575 & \cellcolor[HTML]{FF4D4D}118/582 & \cellcolor[HTML]{FF4D4D}146/554 & \cellcolor[HTML]{FF4D4D}166/535 & \cellcolor[HTML]{55FF00}404/297 & \cellcolor[HTML]{FF4D4D}240/461 & \cellcolor[HTML]{FF4D4D}324/376 & \cellcolor[HTML]{55FF00}396/305 &  \\ \hline
\end{tabular}
\end{sidewaystable}
\twocolumn
}

{
\onecolumn
\begin{sidewaystable}[tpb]
\setlength{\tabcolsep}{2pt}
\vskip 7.0in
\caption{Accuracy of each learner on each datasets. Datasets are indexed from 1 to 70 as follow: (1) MFeatFou, (2) Adult, (3) Vehicule, (4) DressesS, (5) MFeat, (6) SteelPlatesF, (7) Adult, (8) BloodTrans, (9) IntAds, (10) PC1, (11) Nomao, (12) Cylinder, (13) GestureSeg, (14) Madelon, (15) Numerai 28.6, (16) RobotNav, (17) BankMark, (18) CNAE9, (19) KC1, (20) Phishing, (21) CMC, (22) Pen Digits, (23) JChess2PCs, (24) Segment, (25) DNA, (26) MFeatK, (27) OzoneL8, (28) Splice, (29) Wilt, (30) Letter, (31) Semeion, (32) Churn, (33) CreditG, (34) FOTheorem, (35) Mice Protein, (36) Vowel, (37) Balance Scale, (38) ClimateC, (39) JM1, (40) PC4, (41) Spambase, (42) Phoneme, (43) Diabetes, (44) MFeatF, (45) Bioresponce, (46) ILPD, (47) KRvsKP, (48) Satimage, (49) Car, (50) CreditA, (51) AnalcatdataD, (52) Eucalyptus, (53) MFeat Zernike, (54) TicTacToe, (55) Analcatdata, (56) Eletricity, (57) MFeat Pixel, (58) Texture, (59) Iris, (60) Beast W, (61) Isolet, (62) PC3, (63) Sick, (64) Opt Digits, (65) WDBC, (66) Banknote, (67) Har, (68) Connect4, (69) KC2, (70) GSarBD.\label{tab:accuracies}}
\footnotesize
\begin{tabular}{l|ll|llllllllllllllllllllllllllllll}
\hline Learner & Med.Rank & Avg.Rank & 1 & 2 & 3 & 4 & 5 & 6 & 7 & 8 & 9 & 10 & 11 & 12 & 13 & 14 & 15 & 16 & 17 & 18 & 19 & 20 & 21 & 22 & 23 & 24 & 25 & 26 & 27 & 28 & 29 & 30 \\ \hline
YDF Autotuned (opt loss) & 4 & 4.6 & .842 & .870 & .797 & .618 & .729 & .796 & .868 & .778 & .975 & .938 & .968 & .772 & .705 & .828 & .520 & .997 & .907 & .925 & .868 & .971 & .543 & .994 & .947 & .982 & .961 & .970 & .946 & .967 & .985 & .975 \\
YDF Autotuned (opt acc) & 4 & 4.9 & .834 & .873 & .804 & .604 & .717 & .798 & .872 & .779 & .979 & .941 & .973 & .772 & .701 & .835 & .519 & .997 & .909 & .929 & .869 & .971 & .543 & .995 & .954 & .986 & .963 & .967 & .943 & .966 & .984 & .973 \\
LGBM Autotuned (opt loss) & 5 & 5.4 & .834 & .876 & .770 & .592 & .724 & .801 & .874 & .761 & .977 & .939 & .974 & .793 & .668 & .851 & .519 & .996 & .910 & .945 & .856 & .972 & .559 & .992 & .871 & .986 & .964 & .961 & .942 & .964 & .984 & .970 \\
YDF GBT (benchmark hp) & 5 & 5.7 & .843 & .874 & .792 & .564 & .711 & .803 & .872 & .777 & .977 & .938 & .972 & .776 & .692 & .806 & .519 & .996 & .908 & .932 & .860 & .972 & .554 & .994 & .911 & .985 & .962 & .965 & .944 & .965 & .985 & .973 \\
LGBM Autotuned (opt acc) & 5.75 & 5.9 & .832 & .875 & .773 & .582 & .723 & .803 & .874 & .767 & .977 & .940 & .973 & .791 & .676 & .841 & .519 & .996 & .908 & .946 & .856 & .971 & .553 & .992 & .871 & .987 & .963 & .961 & .944 & .964 & .984 & .970 \\
SKLean RF (default hp) & 6 & 6.7 & .834 & .854 & .748 & .606 & .701 & .795 & .854 & .739 & .979 & .940 & .970 & .802 & .683 & .737 & .513 & .995 & .906 & .933 & .867 & .974 & .526 & .992 & .783 & .981 & .957 & .966 & .944 & .969 & .983 & .969 \\
YDF RF (benchmark hp) & 7.5 & 6.8 & .844 & .871 & .764 & .618 & .709 & .803 & .869 & .769 & .974 & .938 & .968 & .774 & .669 & .823 & .520 & .995 & .907 & .859 & .871 & .965 & .552 & .994 & .838 & .980 & .946 & .953 & .944 & .967 & .985 & .964 \\
SKLearn Autotuned & 7 & 7.1 & .833 & .865 & .747 & .606 & .701 & .787 & .864 & .790 & .977 & .934 & .970 & .800 & .682 & .737 & .519 & .995 & .905 & .927 & .864 & .974 & .553 & .992 & .828 & .981 & .956 & .967 & .943 & .968 & .982 & .968 \\
LGBM GBT (default hp) & 8 & 7.9 & .833 & .874 & .760 & .588 & .706 & .798 & .873 & .765 & .976 & .941 & .972 & .789 & .662 & .821 & .517 & .996 & .909 & .850 & .857 & .971 & .555 & .990 & .873 & .986 & .961 & .960 & .940 & .964 & .984 & .966 \\
YDF RF (default hp) & 10 & 9.2 & .839 & .867 & .752 & .598 & .703 & .782 & .866 & .779 & .970 & .935 & .967 & .706 & .652 & .732 & .518 & .995 & .908 & .847 & .869 & .964 & .550 & .989 & .823 & .978 & .955 & .957 & .941 & .969 & .984 & .953 \\
YDF GBT (default hp) & 10 & 9.5 & .808 & .874 & .760 & .572 & .697 & .786 & .873 & .767 & .974 & .941 & .972 & .737 & .645 & .809 & .520 & .997 & .909 & .912 & .856 & .971 & .545 & .989 & .869 & .981 & .962 & .936 & .941 & .963 & .984 & .957 \\
TF Linear (default hp) & 14 & 11.5 & .814 & .852 & .765 & .588 & .729 & .697 & .852 & .777 & .949 & .931 & .936 & .774 & .473 & .558 & .519 & .701 & .887 & .902 & .856 & .928 & .514 & .945 & .674 & .924 & .936 & .939 & .942 & .942 & .946 & .768 \\
XGB GBT (default hp) & 12 & 11.9 & .810 & .869 & .760 & .602 & .714 & .768 & .867 & .770 & .967 & .939 & .966 & .806 & .613 & .810 & .516 & .994 & .904 & .907 & .853 & .954 & .551 & .985 & .861 & .972 & .947 & .929 & .937 & .954 & .981 & .955 \\
XGB Autotuned (acc) & 14 & 12.7 & .806 & .862 & .719 & .566 & .703 & .753 & .859 & .786 & .965 & .926 & .958 & .748 & .601 & .819 & .518 & .988 & .903 & .866 & .852 & .949 & .547 & .976 & .831 & .966 & .949 & .927 & .935 & .956 & .977 & .910 \\
TF EBT (default hp) & 13.25 & 12.8 & .799 & .859 & .739 & .580 & .714 & .751 & .857 & .775 & .967 & .932 & .937 & .759 & .528 & .738 & .519 & .991 & .900 & .839 & .859 & .940 & .553 & .966 & .768 & .969 & .943 & .900 & .937 & .955 & .982 & .803 \\
XGB Autotuned (opt loss) & 14 & 13.5 & .816 & .862 & .734 & .606 & .698 & .752 & .859 & .767 & .960 & .932 & .959 & .730 & .605 & .818 & .518 & .987 & .900 & .864 & .852 & .947 & .559 & .977 & .834 & .967 & .947 & .925 & .935 & .952 & .974 & .916 \\\hline
Learner & 31 & 32 & 33 & 34 & 35 & 36 & 37 & 38 & 39 & 40 & 41 & 42 & 43 & 44 & 45 & 46 & 47 & 48 & 49 & 50 & 51 & 52 & 53 & 54 & 55 & 56 & 57 & 58 & 59 & 60  \\ \hline
YDF Autotuned (opt loss) & & & .946 & .960 & .763 & .619 & 1.000 & .975 & .966 & .944 & .818 & .914 & .954 & .910 & .758 & .970 & .804 & .696 & .994 & .918 & .997 & .878 & .202 & .601 & .810 & .995 & .994 & .917 & .976 & .995 & .940 & .954  \\
YDF Autotuned (opt acc) & & & .942 & .962 & .750 & .620 & 1.000 & .976 & .954 & .933 & .816 & .914 & .954 & .908 & .776 & .967 & .797 & .698 & .994 & .913 & .994 & .861 & .202 & .626 & .798 & .999 & .993 & .944 & .973 & .994 & .953 & .956  \\
LGBM Autotuned (opt loss) & & & .948 & .958 & .740 & .617 & 1.000 & .948 & .862 & .941 & .816 & .910 & .955 & .900 & .744 & .972 & .807 & .678 & .996 & .916 & .995 & .857 & .208 & .621 & .783 & .997 & .993 & .934 & .975 & .987 & .940 & .961  \\
YDF GBT (benchmark hp) & & & .946 & .964 & .746 & .603 & 1.000 & .969 & .947 & .943 & .816 & .912 & .958 & .899 & .767 & .969 & .792 & .678 & .995 & .917 & .996 & .845 & .201 & .560 & .797 & 1.000 & .991 & .904 & .974 & .993 & .947 & .963  \\
LGBM Autotuned (opt acc) & & & .946 & .958 & .727 & .614 & 1.000 & .938 & .856 & .935 & .813 & .906 & .955 & .897 & .724 & .973 & .800 & .683 & .995 & .917 & .997 & .859 & .210 & .626 & .783 & .990 & .993 & .935 & .974 & .986 & .940 & .957  \\
SKLean RF (default hp) & & & .946 & .959 & .769 & .635 & 1.000 & .981 & .816 & .920 & .819 & .911 & .955 & .916 & .759 & .969 & .810 & .696 & .993 & .920 & .964 & .862 & .203 & .616 & .785 & .991 & .992 & .914 & .980 & .979 & .953 & .969  \\
YDF RF (benchmark hp) & & & .912 & .960 & .772 & .626 & 1.000 & .958 & .904 & .935 & .819 & .911 & .951 & .900 & .763 & .968 & .808 & .695 & .991 & .914 & .968 & .852 & .198 & .644 & .781 & .985 & .979 & .897 & .968 & .989 & .953 & .969  \\
SKLearn Autotuned & & & .944 & .959 & .768 & .632 & 1.000 & .982 & .862 & .919 & .818 & .909 & .955 & .917 & .758 & .970 & .810 & .683 & .993 & .919 & .964 & .855 & .197 & .605 & .783 & .991 & .992 & .914 & .979 & .980 & .940 & .964  \\
LGBM GBT (default hp) & & & .930 & .960 & .745 & .607 & 1.000 & .939 & .858 & .946 & .813 & .907 & .954 & .892 & .753 & .961 & .800 & .686 & .992 & .916 & .990 & .852 & .174 & .617 & .775 & .993 & .982 & .924 & .966 & .985 & .947 & .954  \\
YDF RF (default hp) & & & .927 & .959 & .767 & .619 & 1.000 & .948 & .846 & .932 & .816 & .907 & .949 & .898 & .754 & .962 & .807 & .686 & .991 & .911 & .969 & .855 & .210 & .617 & .771 & .985 & .987 & .900 & .974 & .973 & .940 & .959  \\
YDF GBT (default hp) & & & .909 & .957 & .750 & .598 & .999 & .898 & .830 & .935 & .805 & .901 & .954 & .883 & .759 & .952 & .794 & .688 & .995 & .908 & .990 & .844 & .203 & .579 & .767 & .989 & .980 & .915 & .956 & .981 & .967 & .960  \\
TF Linear (default hp) & & & .915 & .865 & .751 & .473 & .996 & .668 & .866 & .950 & .805 & .908 & .907 & .751 & .772 & .969 & .748 & .715 & .958 & .853 & .910 & .868 & .206 & .576 & .826 & .841 & .995 & .753 & .961 & .984 & .913 & .957  \\
XGB GBT (default hp) & & & .895 & .954 & .750 & .603 & .973 & .872 & .816 & .926 & .809 & .900 & .940 & .873 & .750 & .935 & .778 & .666 & .981 & .905 & .954 & .846 & .193 & .598 & .747 & .957 & .960 & .907 & .944 & .973 & .940 & .944  \\
XGB Autotuned (acc) & & & .893 & .948 & .725 & .586 & .982 & .783 & .840 & .922 & .810 & .899 & .932 & .864 & .742 & .954 & .783 & .708 & .979 & .894 & .944 & .854 & .201 & .565 & .747 & .945 & .964 & .846 & .955 & .958 & .953 & .961  \\
TF EBT (default hp) & & & .843 & .955 & .735 & .536 & 1.000 & .834 & .854 & .937 & .812 & .903 & .936 & .860 & .734 & .935 & .779 & .688 & .972 & .879 & .953 & .864 & .188 & .580 & .740 & .971 & .966 & .785 & .937 & .942 & .920 & .960  \\
XGB Autotuned (opt loss) & & & .882 & .939 & .714 & .586 & .979 & .815 & .837 & .917 & .807 & .889 & .931 & .866 & .751 & .952 & .777 & .698 & .973 & .894 & .935 & .858 & .190 & .556 & .740 & .936 & .955 & .846 & .957 & .957 & .887 & .949  \\ \hline
Learner & & & 61 & 62 & 63 & 64 & 65 & 66 & 67 & 68 & 69 & 70 & & & & & & & & & & & & & & & & & & & &  \\ \hline
YDF Autotuned (opt loss) & & & .963 & .900 & .975 & .985 & .972 & 1.000 & .993 & .861 & .839 & .861 & & & & & & & & & & & & & & & & & & & &  \\
YDF Autotuned (opt acc) & & & .961 & .895 & .978 & .983 & .970 & .999 & .993 & .860 & .833 & .865 & & & & & & & & & & & & & & & & & & & &  \\
LGBM Autotuned (opt loss) & & & .960 & .894 & .972 & .985 & .967 & .993 & .993 & .859 & .843 & .865 & & & & & & & & & & & & & & & & & & & &  \\
YDF GBT (benchmark hp) & & & .959 & .891 & .980 & .983 & .977 & .999 & .992 & .847 & .826 & .870 & & & & & & & & & & & & & & & & & & & &  \\
LGBM Autotuned (opt acc) & & & .960 & .897 & .971 & .985 & .961 & .994 & .993 & .860 & .841 & .866 & & & & & & & & & & & & & & & & & & & &  \\
SKLean RF (default hp) & & & .945 & .906 & .967 & .984 & .958 & .993 & .980 & .818 & .845 & .867 & & & & & & & & & & & & & & & & & & & &  \\
YDF RF (benchmark hp) & & & .939 & .899 & .969 & .979 & .967 & .999 & .981 & .804 & .837 & .871 & & & & & & & & & & & & & & & & & & & &  \\
SKLearn Autotuned & & & .947 & .902 & .964 & .984 & .960 & .993 & .981 & .819 & .837 & .869 & & & & & & & & & & & & & & & & & & & &  \\
LGBM GBT (default hp) & & & .952 & .895 & .974 & .980 & .956 & .992 & .991 & .852 & .841 & .861 & & & & & & & & & & & & & & & & & & & &  \\
YDF RF (default hp) & & & .945 & .900 & .950 & .980 & .958 & .991 & .979 & .802 & .833 & .865 & & & & & & & & & & & & & & & & & & & &  \\
YDF GBT (default hp) & & & .930 & .895 & .963 & .974 & .951 & .992 & .989 & .833 & .828 & .865 & & & & & & & & & & & & & & & & & & & &  \\
TF Linear (default hp) & & & .944 & .892 & .963 & .970 & .984 & .981 & .971 & .649 & .843 & .858 & & & & & & & & & & & & & & & & & & & &  \\
XGB GBT (default hp) & & & .940 & .885 & .966 & .968 & .939 & .983 & .983 & .827 & .816 & .841 & & & & & & & & & & & & & & & & & & & &  \\
XGB Autotuned (acc) & & & .932 & .895 & .965 & .966 & .935 & .969 & .970 & .753 & .837 & .845 & & & & & & & & & & & & & & & & & & & &  \\
TF EBT (default hp) & & & .580 & .889 & .966 & .948 & .960 & .990 & .951 & .709 & .822 & .852 & & & & & & & & & & & & & & & & & & & &  \\
XGB Autotuned (opt loss) & & & .932 & .889 & .961 & .965 & .912 & .931 & .971 & .773 & .824 & .837 & & & & & & & & & & & & & & & & & & & & 
\end{tabular}
\end{sidewaystable}
\twocolumn
}

{

\begin{table*}[tpb]
\caption{Name and size of the datasets.}
\label{tab:dataset_stats}
\footnotesize
\begin{tabular}{@{}lrrrr@{}}
\toprule
Dataset &
  \multicolumn{1}{l}{Examples} &
  \multicolumn{1}{l}{Features} &
  \multicolumn{1}{l}{Categorical features} &
  \multicolumn{1}{l}{Numerical features} \\ \midrule
Adult                               & 48842 & 14   & 8  & 6    \\
Adult v2                            & 32561 & 14   & 8  & 6    \\
Analcatdata\_Authorship             & 841   & 70   & 0  & 70   \\
AnalcatData\_Dmft                   & 797   & 4    & 2  & 2    \\
Balance\_Scale                      & 625   & 4    & 0  & 4    \\
Bank\_Marketing                     & 45211 & 16   & 9  & 7    \\
Banknote\_Authentication            & 1372  & 4    & 0  & 4    \\
Beast\_W                            & 699   & 9    & 1  & 8    \\
Bioresponce                         & 3751  & 1776 & 0  & 1776 \\
Blood\_Transfusion\_Service\_Center & 748   & 4    & 0  & 4    \\
Car                                 & 1728  & 6    & 6  & 0    \\
Churn                               & 5000  & 20   & 0  & 20   \\
Climate\_Model\_Simulation\_Crashes & 540   & 20   & 0  & 20   \\
CMC                                 & 1473  & 9    & 0  & 9    \\
CNAE9                               & 1080  & 856  & 0  & 856  \\
Connect4                            & 67557 & 42   & 0  & 42   \\
Credit\_Approval                    & 690   & 15   & 11 & 4    \\
Credit\_G.                          & 1000  & 20   & 13 & 7    \\
Cylinder\_Bands                     & 540   & 39   & 35 & 4    \\
Diabetes                            & 768   & 8    & 0  & 8    \\
DNA                                 & 3186  & 180  & 0  & 180  \\
Dresses\_Sales                      & 500   & 12   & 11 & 1    \\
Eletricity                          & 45312 & 8    & 0  & 8    \\
Eucalyptus                          & 736   & 19   & 14 & 5    \\
First\_Order\_Theorem\_Proving      & 6118  & 51   & 0  & 51   \\
Gesture\_Phase\_Segmentation\_Preocessed &
  9873 &
  32 &
  0 &
  32 \\
GSar\_Bio\_Deg                      & 1055  & 41   & 0  & 41   \\
Har                                 & 10299 & 561  & 0  & 561  \\
ILPD                                & 583   & 10   & 1  & 9    \\
Internet\_Advertisements            & 3279  & 1558 & 0  & 1558 \\
Iris                                & 150   & 4    & 0  & 4    \\
Isolet                              & 7797  & 617  & 0  & 617  \\
JM1                                 & 10885 & 21   & 5  & 16   \\
Jungle\_Chess\_2PCs                 & 44819 & 6    & 0  & 6    \\
KC1                                 & 2109  & 21   & 0  & 21   \\
KC2                                 & 522   & 21   & 0  & 21   \\
KR\_vs\_KP                          & 3196  & 36   & 36 & 0    \\
Letter                              & 20000 & 16   & 0  & 16   \\
Madelon                             & 2600  & 500  & 0  & 500  \\
MFeat\_Factors                      & 2000  & 216  & 0  & 216  \\
MFeat\_Fourier                      & 2000  & 76   & 0  & 76   \\
MFeat\_Karhunen                     & 2000  & 64   & 0  & 64   \\
MFeat\_Morphological                & 2000  & 6    & 0  & 6    \\
MFeat\_Pixel                        & 2000  & 240  & 0  & 240  \\
MFeat\_Zernike                      & 2000  & 47   & 0  & 47   \\
Mice\_Protein                       & 1080  & 81   & 53 & 28   \\
Nomao                               & 34465 & 118  & 0  & 118  \\
Numerai\_28.6                       & 96320 & 21   & 0  & 21   \\
Opt\_Digits                         & 5620  & 64   & 0  & 64   \\
Ozone\_Level\_8h                    & 2534  & 72   & 0  & 72   \\
PC1                                 & 1109  & 21   & 0  & 21   \\
PC3                                 & 1563  & 37   & 0  & 37   \\
PC4                                 & 1458  & 37   & 0  & 37   \\
Pen\_Digits                         & 10992 & 16   & 0  & 16   \\
Phishing\_Websites                  & 11055 & 30   & 0  & 30   \\
Phoneme                             & 5404  & 5    & 0  & 5    \\
Satimage                            & 6430  & 36   & 0  & 36   \\
Segment                             & 2310  & 19   & 0  & 19   \\
Semeion                             & 1593  & 256  & 0  & 256  \\
Sick                                & 3772  & 29   & 29 & 0    \\
Spambase                            & 4601  & 57   & 0  & 57   \\
Splice                              & 3190  & 61   & 61 & 0    \\
Steel\_Plates\_Fault                & 1941  & 27   & 0  & 27   \\
Texture                             & 5500  & 40   & 0  & 40   \\
Tic\_Tac\_Toe                       & 958   & 9    & 9  & 0    \\
Vehicule                            & 846   & 18   & 0  & 18   \\
Vowel                               & 990   & 12   & 2  & 10   \\
Wall\_Robot\_Navigation             & 5456  & 24   & 0  & 24   \\
WDBC                                & 569   & 30   & 0  & 30   \\
Wilt                                & 4839  & 5    & 0  & 5    \\ \bottomrule
\end{tabular}
\end{table*}
}

{
\onecolumn
\begin{sidewaystable}[tpb]
\setlength{\tabcolsep}{2pt}
\vskip 7.0in
\caption{Training time in seconds, averaged over the 10 fold cross-validation, for each learner and each dataset. Datasets are indexed similarly as tab.~\ref{tab:accuracies}. \label{tab:training_time}}
\footnotesize
\begin{tabular}{l|l|llllllllllllllllllllllll}
\hline
Learner & Avg. & 1 & 2 & 2 & 3 & 4 & 5 & 6 & 8 & 9 & 10 & 11 & 12 & 13 & 14 & 15 & 16 & 17 & 18 & 19 & 20 & 22 & 23 & 24 & 25 \\ \hline
YDF GBT (benchmark hp) & 28.74 & 28.13 & 31.69 & 3.92 & 6.29 & 2.95 & 9.60 & 8.75 & 4.79 & 56.94 & 5.25 & 57.92 & 3.42 & 83.13 & 31.61 & 25.96 & 27.95 & 22.84 & 60.24 & 3.58 & 22.45 & 34.49 & 128.10 & 7.76 & 22.06 \\
SKLearn RF (default) & 8.26 & 6.49 & 16.50 & 4.66 & 3.68 & 3.72 & 4.71 & 2.77 & 3.92 & 11.78 & 3.69 & 14.56 & 4.79 & 7.19 & 9.06 & 44.38 & 3.97 & 12.52 & 4.08 & 3.76 & 4.36 & 7.93 & 11.82 & 2.74 & 5.25 \\
YDF TRF (benchmark hp) & 19.81 & 12.22 & 10.63 & 2.73 & 2.76 & 2.57 & 3.05 & 5.31 & 2.52 & 112.10 & 2.77 & 69.26 & 2.70 & 26.74 & 43.90 & 93.70 & 6.86 & 13.08 & 27.63 & 3.19 & 4.75 & 6.15 & 7.80 & 3.47 & 9.51 \\
LGBM GBT (default) & 1.22 & 3.78 & 0.94 & 0.05 & 0.12 & 0.04 & 0.15 & 0.65 & 0.03 & 0.27 & 0.06 & 4.01 & 0.05 & 5.45 & 1.50 & 0.92 & 1.06 & 0.83 & 0.16 & 0.07 & 0.43 & 1.66 & 5.74 & 0.79 & 0.47 \\
YDF RF (default) & 4.60 & 3.88 & 5.27 & 2.46 & 2.46 & 2.33 & 2.70 & 2.83 & 2.37 & 5.22 & 2.52 & 8.65 & 2.49 & 8.44 & 4.68 & 19.46 & 3.17 & 5.11 & 4.99 & 2.52 & 3.17 & 3.40 & 5.92 & 2.68 & 3.66 \\
YDF GBT (default) & 20.61 & 32.99 & 13.56 & 3.40 & 4.91 & 4.03 & 7.87 & 8.94 & 2.77 & 39.38 & 3.29 & 72.21 & 3.25 & 56.64 & 30.01 & 9.47 & 14.79 & 23.50 & 50.79 & 3.33 & 17.98 & 29.24 & 34.34 & 8.01 & 19.97 \\
TF Linear (dedault) & 68.15 & 23.10 & 55.12 & 6.93 & 8.52 & 9.01 & 6.40 & 12.84 & 5.94 & 564.50 & 11.76 & 118.00 & 15.94 & 27.34 & 176.10 & 107.40 & 14.46 & 60.73 & 210.20 & 14.61 & 23.14 & 18.15 & 53.38 & 12.65 & 79.39 \\
XGB GBT (default) & 6.70 & 7.89 & 17.26 & 0.07 & 0.23 & 0.03 & 0.37 & 1.56 & 0.01 & 10.13 & 0.09 & 23.99 & 0.34 & 35.00 & 2.76 & 1.95 & 2.43 & 6.71 & 16.55 & 0.10 & 2.28 & 8.89 & 17.49 & 1.23 & 3.34 \\
TF EBT (default) & 91.45 & 112.40 & 62.17 & 12.62 & 11.73 & 7.05 & 24.03 & 31.60 & 7.62 & 216.70 & 7.50 & 96.24 & 13.08 & 68.11 & 77.72 & 136.70 & 26.35 & 62.48 & 857.40 & 10.06 & 21.87 & 131.40 & 81.73 & 35.04 & 83.15 \\ \hline
 &  & 26 & 27 & 28 & 29 & 30 & 31 & 32 & 33 & 34 & 35 & 36 & 37 & 38 & 39 & 40 & 41 & 42 & 43 & 44 & 45 & 46 & 47 & 48 & 49 \\\hline
YDF GBT (benchmark hp) &  & 24.47 & 4.88 & 7.54 & 4.56 & 251.00 & 38.31 & 7.27 & 3.44 & 21.27 & 29.97 & 8.93 & 5.02 & 3.42 & 8.63 & 4.89 & 9.55 & 7.10 & 2.98 & 37.83 & 53.99 & 3.00 & 10.53 & 21.36 & 14.59 \\
SKLearn RF (default) &  & 4.92 & 3.81 & 5.08 & 3.74 & 9.33 & 3.51 & 4.46 & 5.00 & 5.50 & 3.22 & 4.03 & 2.90 & 2.31 & 9.55 & 4.94 & 4.96 & 5.45 & 2.58 & 4.66 & 11.80 & 3.44 & 3.38 & 3.49 & 4.82 \\
YDF TRF (benchmark hp) &  & 10.03 & 4.53 & 4.35 & 2.79 & 13.93 & 13.00 & 4.63 & 2.70 & 21.33 & 3.54 & 3.20 & 2.41 & 2.60 & 25.21 & 3.08 & 4.99 & 4.99 & 2.58 & 17.40 & 370.80 & 2.46 & 2.62 & 11.58 & 2.40 \\
LGBM GBT (default) &  & 5.36 & 0.28 & 0.59 & 0.07 & 8.82 & 1.35 & 0.18 & 0.04 & 2.38 & 1.36 & 0.96 & 0.06 & 0.06 & 0.19 & 0.09 & 0.33 & 0.13 & 0.03 & 8.97 & 2.17 & 0.03 & 0.15 & 0.99 & 0.23 \\
YDF RF (default) &  & 3.55 & 2.76 & 2.78 & 2.47 & 4.75 & 3.64 & 2.89 & 2.42 & 5.27 & 2.83 & 2.70 & 2.32 & 2.47 & 3.51 & 2.42 & 3.15 & 2.77 & 2.53 & 3.59 & 12.27 & 2.33 & 2.56 & 3.28 & 2.30 \\
YDF GBT (default) &  & 28.73 & 4.04 & 6.76 & 3.21 & 75.54 & 53.58 & 4.09 & 3.33 & 24.85 & 52.74 & 8.72 & 5.58 & 3.00 & 3.50 & 4.84 & 11.27 & 4.54 & 2.89 & 31.44 & 30.98 & 3.37 & 5.44 & 24.83 & 9.59 \\
TF Linear (dedault) &  & 24.08 & 33.50 & 24.81 & 11.31 & 28.38 & 76.31 & 16.39 & 11.85 & 27.16 & 14.12 & 7.21 & 6.15 & 9.80 & 23.58 & 18.33 & 37.45 & 11.03 & 6.03 & 73.92 & 610.40 & 7.70 & 20.07 & 20.03 & 6.68 \\
XGB GBT (default) &  & 8.11 & 0.42 & 5.92 & 0.17 & 73.57 & 11.92 & 0.32 & 0.12 & 12.57 & 2.08 & 1.52 & 0.07 & 0.05 & 7.84 & 0.14 & 0.83 & 0.31 & 0.02 & 12.75 & 9.85 & 0.02 & 1.64 & 4.68 & 0.70 \\
TF EBT (default) &  & 102.80 & 20.86 & 22.34 & 10.84 & 1350.00 & 462.00 & 12.55 & 8.46 & 101.20 & 31.27 & 29.06 & 7.31 & 8.62 & 43.20 & 12.28 & 19.50 & 14.19 & 6.69 & 340.40 & 345.00 & 6.02 & 13.11 & 76.87 & 7.85 \\ \hline
 &  & 50 & 51 & 52 & 53 & 53 & 55 & 57 & 57 & 58 & 60 & 61 & 62 & 63 & 64 & 65 & 66 & 67 & 68 & 69 & 70 &  &  &  &  \\ \hline
YDF GBT (benchmark hp) &  & 3.45 & 3.35 & 5.83 & 17.24 & 4.80 & 21.61 & 75.14 & 42.23 & 53.75 & 3.74 & 697.50 & 3.59 & 2.93 & 34.22 & 6.12 & 6.37 & 238.80 & 216.00 & 3.01 & 5.28 &  &  &  &  \\
SKLearn RF (default) &  & 3.12 & 4.82 & 3.18 & 6.49 & 2.53 & 3.60 & 13.17 & 1.96 & 7.91 & 5.38 & 21.67 & 3.01 & 7.06 & 8.23 & 4.40 & 3.62 & 28.78 & 23.97 & 5.50 & 2.80 &  &  &  &  \\
YDF TRF (benchmark hp) &  & 2.54 & 2.42 & 3.77 & 8.92 & 2.47 & 3.17 & 12.62 & 9.28 & 10.17 & 2.40 & 631.90 & 3.35 & 11.83 & 7.83 & 2.47 & 2.43 & 255.80 & 39.40 & 2.39 & 2.96 &  &  &  &  \\
LGBM GBT (default) &  & 0.05 & 0.04 & 0.08 & 2.26 & 0.07 & 0.43 & 3.11 & 2.56 & 4.86 & 0.04 & 184.60 & 0.08 & 0.08 & 1.23 & 0.11 & 0.11 & 55.75 & 14.18 & 0.03 & 0.09 &  &  &  &  \\
YDF RF (default) &  & 2.40 & 2.40 & 2.47 & 5.38 & 2.34 & 2.45 & 5.99 & 3.29 & 4.08 & 2.39 & 27.40 & 2.51 & 2.40 & 3.40 & 2.40 & 2.34 & 16.45 & 13.06 & 2.37 & 2.45 &  &  &  &  \\
YDF GBT (default) &  & 3.06 & 3.09 & 4.35 & 10.84 & 6.57 & 10.63 & 27.73 & 32.52 & 36.55 & 3.15 & 872.40 & 5.02 & 3.17 & 19.32 & 3.37 & 8.72 & 267.00 & 139.70 & 2.70 & 4.66 &  &  &  &  \\
TF Linear (dedault) &  & 9.29 & 5.22 & 8.91 & 16.38 & 6.87 & 23.69 & 47.12 & 71.48 & 22.76 & 8.26 & 307.40 & 13.79 & 24.14 & 35.80 & 10.88 & 6.83 & 280.90 & 92.53 & 10.89 & 15.16 &  &  &  &  \\
XGB GBT (default) &  & 0.09 & 0.06 & 1.49 & 5.28 & 0.14 & 0.51 & 11.52 & 10.74 & 12.93 & 0.05 & 627.10 & 0.13 & 6.17 & 10.07 & 0.11 & 0.10 & 245.00 & 161.00 & 0.03 & 0.11 &  &  &  &  \\
TF EBT (default) &  & 8.30 & 9.99 & 13.05 & 91.09 & 6.81 & 30.65 & 48.43 & 477.90 & 141.30 & 7.51 & 6862.00 & 10.88 & 24.71 & 181.00 & 8.35 & 7.26 & 988.70 & 236.90 & 7.63 & 10.20 &  &  &  &  \\ \hline
\end{tabular}
\end{sidewaystable}
\twocolumn
}

{
\onecolumn
\begin{sidewaystable}[tpb]
\setlength{\tabcolsep}{2pt}
\vskip 7.0in
\caption{Inference time of the test set, averaged over the 10 fold cross-validation, for each learner and each dataset. Datasets are indexed similarly as tab.~\ref{tab:accuracies}. \label{tab:inference_time}}
\footnotesize
\begin{tabular}{l|l|lllllllllllllllllllllllll}
\hline
Learner & Avg. & 1 & 2 & 2 & 3 & 4 & 5 & 6 & 8 & 9 & 10 & 11 & 12 & 13 & 14 & 15 & 16 & 17 & 18 & 19 & 20 & 22 & 23 & 24 & 25 & 26 \\ \hline
YDF GBT (benchmark hp) & 0.090 & 0.028 & 0.099 & 0.004 & 0.011 & 0.001 & 0.015 & 0.016 & 0.002 & 0.142 & 0.002 & 0.118 & 0.005 & 0.184 & 0.025 & 0.020 & 0.064 & 0.052 & 0.108 & 0.002 & 0.056 & 0.135 & 1.044 & 0.019 & 0.032 & 0.054 \\
SKLearn RF (default) & 0.260 & 0.232 & 0.281 & 0.154 & 0.166 & 0.173 & 0.225 & 0.292 & 0.163 & 0.324 & 0.172 & 0.259 & 0.185 & 0.337 & 0.183 & 0.506 & 0.378 & 0.354 & 0.192 & 0.160 & 0.206 & 0.308 & 0.514 & 0.220 & 0.316 & 0.197 \\
YDF RF (benchmark hp) & 0.767 & 0.145 & 2.271 & 0.163 & 0.063 & 0.028 & 0.161 & 0.148 & 0.051 & 0.985 & 0.035 & 1.318 & 0.025 & 1.144 & 0.420 & 5.258 & 0.076 & 2.076 & 0.425 & 0.094 & 0.421 & 0.313 & 3.175 & 0.052 & 0.196 & 0.129 \\
LGBM GBT (default) & 0.043 & 0.015 & 0.037 & 0.004 & 0.004 & 0.003 & 0.010 & 0.010 & 0.003 & 0.020 & 0.003 & 0.068 & 0.003 & 0.096 & 0.009 & 0.020 & 0.021 & 0.026 & 0.015 & 0.004 & 0.022 & 0.071 & 0.562 & 0.019 & 0.011 & 0.023 \\
YDF RF (default) & 0.434 & 0.145 & 1.353 & 0.121 & 0.045 & 0.026 & 0.157 & 0.120 & 0.041 & 0.034 & 0.022 & 0.652 & 0.014 & 0.856 & 0.117 & 2.400 & 0.062 & 1.276 & 0.046 & 0.089 & 0.224 & 0.344 & 2.347 & 0.038 & 0.210 & 0.120 \\
YDF GBT (default) & 0.036 & 0.012 & 0.017 & 0.003 & 0.004 & 0.002 & 0.007 & 0.009 & 0.000 & 0.003 & 0.001 & 0.034 & 0.002 & 0.084 & 0.004 & 0.008 & 0.024 & 0.015 & 0.012 & 0.001 & 0.011 & 0.085 & 0.519 & 0.011 & 0.012 & 0.022 \\
TF Linear (dedault) & 10.013 & 4.710 & 1.236 & 0.757 & 1.469 & 1.371 & 0.547 & 1.918 & 0.441 & 100.200 & 1.240 & 8.541 & 2.639 & 2.257 & 35.440 & 1.623 & 1.524 & 1.251 & 59.670 & 1.550 & 2.399 & 1.224 & 0.625 & 1.457 & 12.300 & 3.930 \\
XGB GBT (default) & 0.010 & 0.005 & 0.010 & 0.002 & 0.003 & 0.000 & 0.004 & 0.004 & 0.000 & 0.003 & 0.001 & 0.014 & 0.001 & 0.026 & 0.001 & 0.003 & 0.004 & 0.007 & 0.005 & 0.000 & 0.005 & 0.030 & 0.119 & 0.005 & 0.004 & 0.006 \\
TF EBT (default) & 2.102 & 1.087 & 0.702 & 0.408 & 0.495 & 0.591 & 0.371 & 0.549 & 0.367 & 19.470 & 0.555 & 1.609 & 1.667 & 0.652 & 6.620 & 0.688 & 0.531 & 0.662 & 9.656 & 0.569 & 0.671 & 0.432 & 0.409 & 0.472 & 2.027 & 1.285 \\ \hline
 &  & 27 & 28 & 29 & 30 & 31 & 32 & 33 & 34 & 35 & 36 & 37 & 38 & 39 & 40 & 41 & 42 & 43 & 44 & 45 & 46 & 47 & 48 & 49 & 50 & 51 \\ \hline
YDF GBT (benchmark hp) &  & 0.003 & 0.006 & 0.006 & 1.339 & 0.073 & 0.009 & 0.001 & 0.050 & 0.056 & 0.046 & 0.008 & 0.001 & 0.011 & 0.005 & 0.013 & 0.020 & 0.001 & 0.048 & 0.073 & 0.001 & 0.005 & 0.037 & 0.029 & 0.001 & 0.003 \\
SKLearn RF (default) &  & 0.296 & 0.239 & 0.178 & 0.436 & 0.257 & 0.384 & 0.159 & 0.292 & 0.193 & 0.235 & 0.199 & 0.189 & 0.223 & 0.192 & 0.310 & 0.323 & 0.133 & 0.309 & 0.439 & 0.151 & 0.224 & 0.303 & 0.230 & 0.172 & 0.163 \\
YDF RF (benchmark hp) &  & 0.061 & 0.146 & 0.050 & 2.020 & 0.230 & 0.164 & 0.069 & 0.721 & 0.027 & 0.077 & 0.026 & 0.011 & 0.744 & 0.043 & 0.243 & 0.235 & 0.032 & 0.124 & 3.085 & 0.032 & 0.048 & 0.286 & 0.033 & 0.025 & 0.060 \\
LGBM GBT (default) &  & 0.004 & 0.013 & 0.005 & 0.771 & 0.016 & 0.006 & 0.003 & 0.034 & 0.017 & 0.015 & 0.004 & 0.009 & 0.005 & 0.004 & 0.007 & 0.008 & 0.003 & 0.026 & 0.025 & 0.003 & 0.006 & 0.031 & 0.022 & 0.005 & 0.004 \\
YDF RF (default) &  & 0.040 & 0.140 & 0.025 & 1.462 & 0.165 & 0.094 & 0.049 & 0.554 & 0.025 & 0.059 & 0.029 & 0.007 & 0.532 & 0.032 & 0.102 & 0.171 & 0.034 & 0.125 & 0.242 & 0.026 & 0.062 & 0.236 & 0.065 & 0.040 & 0.062 \\
YDF GBT (default) &  & 0.001 & 0.011 & 0.001 & 0.639 & 0.033 & 0.002 & 0.001 & 0.026 & 0.060 & 0.013 & 0.003 & 0.001 & 0.002 & 0.002 & 0.003 & 0.004 & 0.001 & 0.018 & 0.006 & 0.001 & 0.008 & 0.020 & 0.027 & 0.001 & 0.002 \\
TF Linear (dedault) &  & 4.563 & 3.803 & 0.561 & 1.182 & 15.110 & 1.426 & 1.615 & 3.343 & 2.902 & 1.002 & 0.567 & 1.271 & 1.503 & 2.470 & 3.537 & 0.563 & 0.744 & 13.700 & 108.400 & 0.885 & 2.990 & 2.301 & 0.630 & 1.432 & 0.504 \\
XGB GBT (default) &  & 0.001 & 0.004 & 0.001 & 0.279 & 0.007 & 0.001 & 0.001 & 0.008 & 0.003 & 0.005 & 0.003 & 0.001 & 0.004 & 0.001 & 0.002 & 0.001 & 0.001 & 0.006 & 0.003 & 0.001 & 0.008 & 0.009 & 0.004 & 0.001 & 0.003 \\
TF EBT (default) &  & 1.379 & 1.784 & 0.326 & 0.580 & 3.175 & 0.624 & 0.626 & 1.056 & 0.886 & 0.454 & 0.347 & 0.491 & 1.575 & 0.626 & 0.963 & 0.334 & 0.366 & 3.641 & 22.590 & 0.502 & 1.344 & 0.674 & 0.401 & 0.760 & 0.334 \\ \hline
 &  & 52 & 53 & 53 & 55 & 57 & 57 & 58 & 60 & 61 & 62 & 63 & 64 & 65 & 66 & 67 & 68 & 69 & 70 &  &  &  &  &  &  &  \\ \hline
YDF GBT (benchmark hp) &  & 0.009 & 0.022 & 0.007 & 0.034 & 0.270 & 0.079 & 0.123 & 0.003 & 0.909 & 0.003 & 0.003 & 0.085 & 0.004 & 0.017 & 0.204 & 1.520 & 0.002 & 0.003 &  &  &  &  &  &  &  \\
SKLearn RF (default) &  & 0.297 & 0.274 & 0.152 & 0.142 & 0.353 & 0.206 & 0.280 & 0.122 & 0.240 & 0.150 & 0.246 & 0.386 & 0.142 & 0.305 & 0.306 & 0.397 & 0.165 & 0.143 &  &  &  &  &  &  &  \\
YDF RF (benchmark hp) &  & 0.065 & 0.146 & 0.044 & 0.023 & 2.025 & 0.147 & 0.171 & 0.011 & 1.648 & 0.060 & 0.063 & 0.289 & 0.009 & 0.013 & 0.643 & 7.489 & 0.021 & 0.051 &  &  &  &  &  &  &  \\
LGBM GBT (default) &  & 0.004 & 0.013 & 0.005 & 0.008 & 0.244 & 0.019 & 0.056 & 0.003 & 0.283 & 0.003 & 0.005 & 0.039 & 0.003 & 0.004 & 0.076 & 1.242 & 0.003 & 0.003 &  &  &  &  &  &  &  \\
YDF RF (default) &  & 0.067 & 0.126 & 0.032 & 0.022 & 1.160 & 0.129 & 0.175 & 0.006 & 1.153 & 0.038 & 0.021 & 0.252 & 0.004 & 0.006 & 0.545 & 3.133 & 0.011 & 0.034 &  &  &  &  &  &  &  \\
YDF GBT (default) &  & 0.006 & 0.011 & 0.005 & 0.012 & 0.034 & 0.033 & 0.049 & 0.003 & 0.331 & 0.001 & 0.003 & 0.054 & 0.001 & 0.003 & 0.116 & 0.534 & 0.000 & 0.001 &  &  &  &  &  &  &  \\
TF Linear (dedault) &  & 1.567 & 3.299 & 0.786 & 4.814 & 1.118 & 13.840 & 2.991 & 0.907 & 40.150 & 2.289 & 2.052 & 3.758 & 1.937 & 0.443 & 32.700 & 3.191 & 1.471 & 2.742 &  &  &  &  &  &  &  \\
XGB GBT (default) &  & 0.003 & 0.005 & 0.001 & 0.003 & 0.044 & 0.006 & 0.010 & 0.001 & 0.057 & 0.000 & 0.002 & 0.013 & 0.000 & 0.001 & 0.029 & 0.227 & 0.001 & 0.001 &  &  &  &  &  &  &  \\
TF EBT (default) &  & 0.892 & 0.664 & 0.424 & 1.247 & 0.464 & 3.674 & 1.038 & 0.408 & 9.157 & 0.807 & 1.917 & 1.041 & 0.588 & 0.335 & 9.433 & 0.924 & 0.469 & 0.686 &  &  &  &  &  &  &  \\ \hline
\end{tabular}
\end{sidewaystable}
\twocolumn
}

\end{document}